\documentclass[sigconf,noncam]{acmart}	

\usepackage{booktabs}
\usepackage{enumitem}
\usepackage{algorithm}
\usepackage{algpseudocode}

\AtBeginDocument{%
  \providecommand\BibTeX{{%
    \normalfont B\kern-0.5em{\scshape i\kern-0.25em b}\kern-0.8em\TeX}}}

\copyrightyear{2023}
\acmYear{2023}
\setcopyright{none}
\renewcommand\footnotetextcopyrightpermission[1]{} % removes footnote with conference 

\begin{document}

\title{Uni-paint: A Unified Framework for Multimodal Image Inpainting with Pretrained Diffusion Model}

\author{Shiyuan Yang}
\affiliation{%
  \institution{City University of Hong Kong}
  \city{Hong Kong SAR}
  \country{China}}

\author{Xiaodong Chen}
\affiliation{%
  \institution{Tianjin University}
  \city{Tianjin}
  \country{China}}

\author{Jing Liao}
\authornote{Corresponding author.}
\affiliation{%
  \institution{City University of Hong Kong}
  \city{Hong Kong SAR}
  \country{China}}

\begin{abstract}
Recently, text-to-image denoising diffusion probabilistic models (DDPMs) have demonstrated impressive image generation capabilities and have also been successfully applied to image inpainting. However, in practice, users often require more control over the inpainting process beyond textual guidance, especially when they want to composite objects with customized appearance, color, shape, and layout. Unfortunately, existing diffusion-based inpainting methods are limited to single-modal guidance and require task-specific training, hindering their cross-modal scalability. To address these limitations, we propose Uni-paint, a unified framework for multimodal inpainting that offers various modes of guidance, including unconditional, text-driven, stroke-driven, exemplar-driven inpainting, as well as a combination of these modes. Furthermore, our Uni-paint is based on pretrained Stable Diffusion and does not require task-specific training on specific datasets, enabling few-shot generalizability to customized images. We have conducted extensive qualitative and quantitative evaluations that show our approach achieves comparable results to existing single-modal methods while offering multimodal inpainting capabilities not available in other methods. Code will be available at \url{https://github.com/ysy31415/unipaint}.
\end{abstract}

\begin{teaserfigure}
  \includegraphics[width=\textwidth]{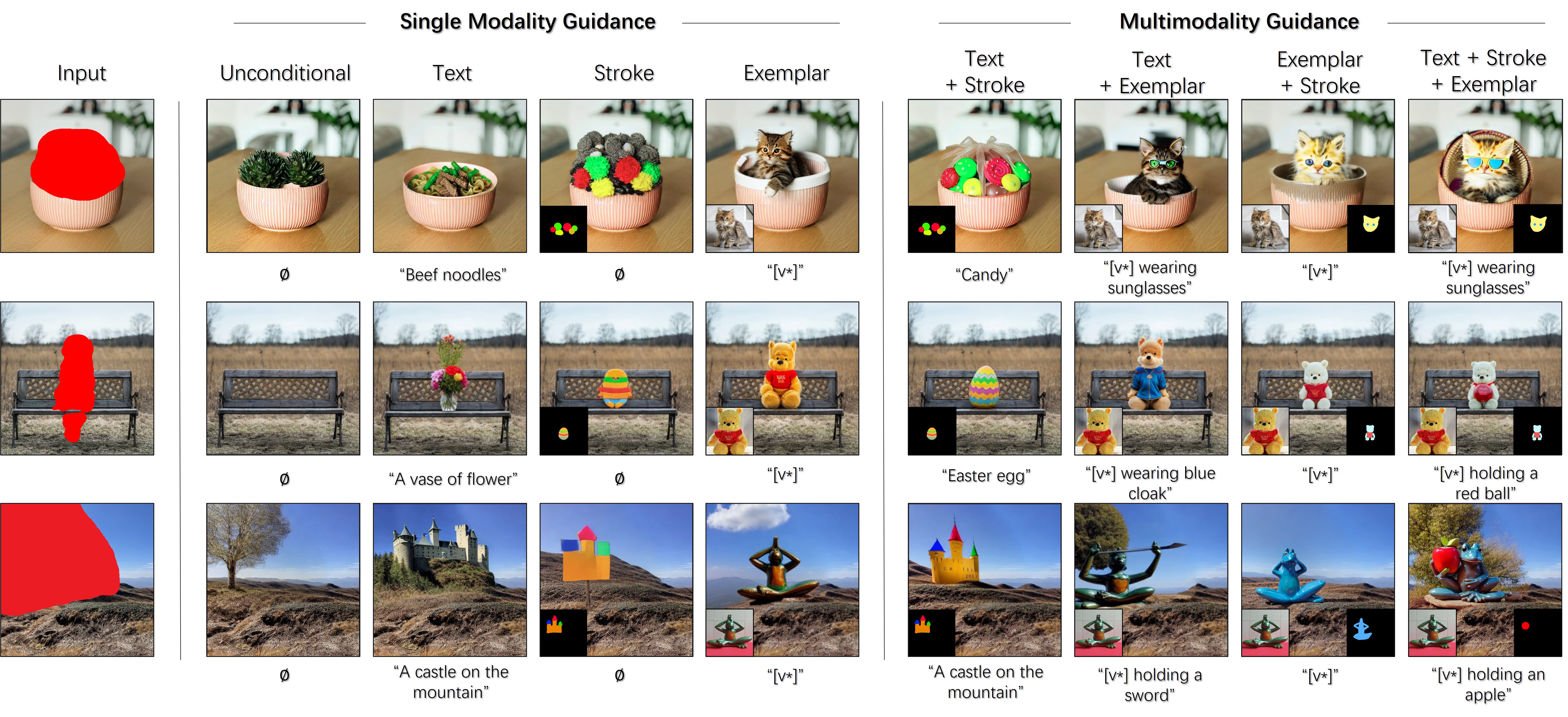}
  \caption{Uni-paint allows users to perform unconditional, text-guided, stroke-guided, exemplar-guided or mix-guided inpainting on a single provided image within one unified framework.}
  \label{fig.teaser}
\end{teaserfigure}

\maketitle
\pagestyle{plain} % removes running headers

%%%%%%%%%%%%%%%%%%%%%%%%%%%%%%%%%%%%%%%%%%%%%%%%%%%%%
\section{Introduction}\label{sec.intro}
%%%%%%%%%%%%%%%%%%%%%%%%%%%%%%%%%%%%%%%%%%%%%%%%%%%%%

Large-scale language-image models such as denoising diffusion probabilistic models (DDPMs) \cite{ddpm, ldm, imagen, dalle2} have recently shown impressive generation quality and domain generizibility surpassing that of GANs. As a promising generative modeling paradigm, diffusion models have also been applied to image inpainting. Current diffusion-based inpainting methods can be divided into two categories: training-based and few-shot methods. Training-based methods involve either training an image-to-image diffusion model \cite{plaette} or modifying a pretrained text-to-image model \cite{glide, ldm, smartbrush, editbench} with additional conditioning (e.g., masked image). While these models have fast inference times, they suffer from several drawbacks. Firstly, large-scale dataset acquisition can be challenging. Secondly, these methods are less scalable for modal extension as they have been specifically designed and trained for a certain modality. On the other hand, few-shot methods directly leverage the powerful generative capability of an off-the-shelf pretrained model through model prior \cite{repaint, bld} or guided sampling \cite{bd}, requiring no additional dataset collection and training. This category offers higher flexibility and scalability, making it preferred in our work.

Despite the success of diffusion models in the inpainting task, their potential has not been fully exploited yet. Current methods, regardless of their category, have limited capabilities for modal extension, supporting only unconditional inpainting \cite{repaint} or single modality guidance (text guidance \cite{ldm, glide, bld}, or exemplar-guidance \cite{pbe}). This lack of flexibility in general usage can be problematic, as a combination of different interaction methods is often required to achieve a satisfactory inpainting result. For example, while textual descriptions can be used to describe high-level semantics to be inpainted, they may struggle to accurately convey the user's intentions for object shape, color, and customized attributes. This can be addressed by providing an additional reference image (exemplar) or drawing rough color strokes. As shown by the cat at upper right corner in Fig.~\ref{fig.teaser}, generating such cat with specific identity, predetermined colors, and gestures is much easier using multiple guidance rather than relying solely on text. Therefore, a framework that enables multimodal conditions for image inpainting is a natural choice, but existing methods do not support it.

In this work, we present Uni-paint, the first unified framework for multimodal image inpainting that supports both unconditional and conditional controls, including text, stroke, exemplar, and a combination of them, as shown in Fig.~\ref{fig.teaser}. To achieve this, we first finetune a pretrained Stable Diffusion model \cite{ldm} unconditionally, requiring it to generate images that are only faithful to the known part of the input image, which we refer to as masked finetuning. Since the Stable Diffusion has been extensively pretrained on large image datasets, it possesses the prior knowledge needed to generate plausible images. Our masked finetuning further enables the model to generate context-plausible content in the unknown region unconditionally by leveraging its learned semantic awareness of the known part. 
%Therefore, unconditional inpainting can be automatically achieved by sampling the finetuned model unconditionally. 
Furthermore, by exploiting the existing conditional interface of a text-to-image diffusion model, conditional inpainting with multiple modalities is also unified in this framework. We identify two types of conditional interfaces: (1) the textual interface, implemented through cross-attention, applicable for semantic guidance like text and exemplar, and (2) the spatial interface, achieved through image blending, suitable for spatial guidance like stroke. These guidance modes can even be combined in the same framework to perform mixed-modal inpainting.

Our Uni-paint is a few-shot method that differs from previous approaches \cite{plaette,glide, ldm} that require training the model on large datasets. Our method only requires finetuning on a single input image, reducing the dependency on data collection and eliminating restrictions to training domain. However, like other few-shot inpainting methods \cite{repaint, bld, bd}, our approach needs to progressively blend the inpainted content with the known regions of the input image during the sampling process to keep the known region untouched. A common issue in blending-based methods is that the inpainted content may overflow the mask boundary and get truncated after blending. 
%This is challenging, as controlling the spatial position of objects is difficult in the generation of diffusion models. 
To address this, we introduce a masked attention control mechanism for cross-attention and self-attention layers of the diffusion model to restrict the scope of the generated content within the unknown area.

Our Uni-paint framework has undergone extensive qualitative and quantitative evaluations that demonstrate its comparable results to existing single-modal methods while offering multimodal inpainting capabilities that are not available in other methods. In summary, our contributions are as follows: 
\begin{itemize}[itemsep=0pt,topsep=1pt,parsep=0pt,leftmargin=*]
\item We propose the first unified framework for multimodal image inpainting based on pretrained diffusion model.
\item We introduce few-shot masked finetuning on a single image with null conditioning, making the inpainting scalable to other modalities and generalizable to customized image inputs.
\item We introduce a masked attention mechanism to alleviate the potential leakage of inpainted content to known areas.
\end{itemize}

%%%%%%%%%%%%%%%%%%%%%%%%%%%%%%%%%%%%%%%%%%%%%%%%%%%%%
\section{Related work} \label{sec.relatedwork}
%%%%%%%%%%%%%%%%%%%%%%%%%%%%%%%%%%%%%%%%%%%%%%%%%%%%%

\noindent \textbf{Image inpainting}. 
Early methods relied on borrowing the low-level texture patches from known regions \cite{pm, criminisi, km}, but struggled with complex semantic scenes like face completion. This problem was not solved until ContextEncoder \cite{ce}, the first GAN-based model was proposed. Subsequent CNN/GAN-based works achieved improved results by incorporating various modules like Partial Convolution \cite{pconv}, Contextual Attention \cite{dfv1}, Fourier Convolution \cite{lama}, etc.
% and strategies. Liu et al. \cite{pconv} proposed Partial Convolution to handle the missing areas, Yu et al. \cite{dfv1} introduced Contextual Attention to boost spatial feature swapping. Suvorov et al. \cite{lama} presented Fourier Convolution for effective global and local information utilization. 
Others works used multi-stage pipelines like edge-guided \cite{ec,fa}, coarse-to-fine \cite{ict,dfv2,imt}, progressive \cite{pgn,prvs}, and recurrent \cite{rfr} networks. The recent RePaint method \cite{repaint} leveraged DDPM priors and repetitive sampling for promising results. However, these unconditional methods do not support user guidance.

%-----------
\noindent \textbf{Text-driven image editing}. % and inpainting}. 
As a user-friendly guidance modality, text-driven editing has been gaining popularity. Early GAN-based method like StyleCLIP \cite{styleclip} achieved human face editing by leveraging pretrained StyleGAN \cite{stylegan} and CLIP \cite{clip}.  
Recently, the rise of large text-to-image diffusion models \cite{ldm, glide, dalle2, imagen} has paved a promising way for high-quality, high-diversity text-driven generative modeling. These works can be roughly divided into three categories: (1) Guided Sampling: by introducing various guidance functions during sampling, such as CLIP guidance \cite{bd}, style guidance \cite{sdg}, edge guidance \cite{sketchdiffusion}, and attention guidance \cite{attnexcite}. (2) Attention Control: Works like Prompt-to-Prompt\cite{p2p} and shape-guided editing \cite{ioattn} manipulate the attention layer for impressive editing results. (3) Training/Finetuning: InstructPix2Pix \cite{instructpix2pix} trained a model on generated image-prompt pairs for fast editing with user instructions. Few-shot editing can also be achieved by finetuning the model \cite{unitune,diffusionclip}, optimizing the text embedding \cite{txtinv, nulltxtinv}, or both \cite{imagic}. 
% del
Text-driven image editing with diffusion models has also been applied to image inpainting by altering a pretrained text-to-image model to enable masked image conditioning \cite{ldm,glide,editbench,smartbrush}, but these methods require large training datasets.
% Text-driven image editing with diffusion models has also been applied to image inpainting. SD-inpaint \cite{ldm}, GLIDE \cite{glide}, EditBench \cite{editbench} and SmartBrush \cite{smartbrush} altered a pretrained text-to-image model to enable masked image conditioning, but they require large training datasets. 
Alternatively, Blended Diffusion \cite{bd,bld} achieved zero-shot inpainting by utilizing background blending. However, the blending strategy does not expose the model to full context and may fail to handle semantic transitions near the hole boundary. In contrast, our method employs masked finetuning and attention control, ensuring global context awareness and improved texture transitions.

%-----------
\noindent \textbf{Exemplar-driven image editing}. % and inpainting}. 
Exemplar-driven image editing is a relatively new topic, which allows users to synthesize customized objects using provided exemplars. Textual inversion \cite{txtinv} captures new concepts from exemplars by optimizing token embedding. DreamBooth \cite{db} generates personalized outputs by finetuning the model on exemplar images. Follow-up works like CustomDiff \cite{customdiff}, SVDiff \cite{svdiff}, and ELITE \cite{elite} further accelerate the process by finetuning the model's cross-attention layer, singular values of the weights, and learning a mapping encoder, respectively.
% These works do not restrict subject's spatial location, but this can be a problem when it comes to inpainting. 
Paint-by-Example \cite{pbe} first achieved exemplar-driven inpainting by trading textual conditioning for image conditioning, leading to a lack of text guidance support. Also, its generalization to unseen objects may also be limited by the training dataset. Our approach mitigates these issues by extending a pretrained text-to-image model for user-specific sample finetuning without modifying the model structure.

%%%%%%%%%%%%%%%%%%%%%%%%%%%%%%%%%%%%%%%%%%%%%%%%%%%%%
\section{Method} \label{sec.method}
%%%%%%%%%%%%%%%%%%%%%%%%%%%%%%%%%%%%%%%%%%%%%%%%%%%%%
\begin{figure*}[t]
\begin{center}
   \includegraphics[width=0.9\linewidth]{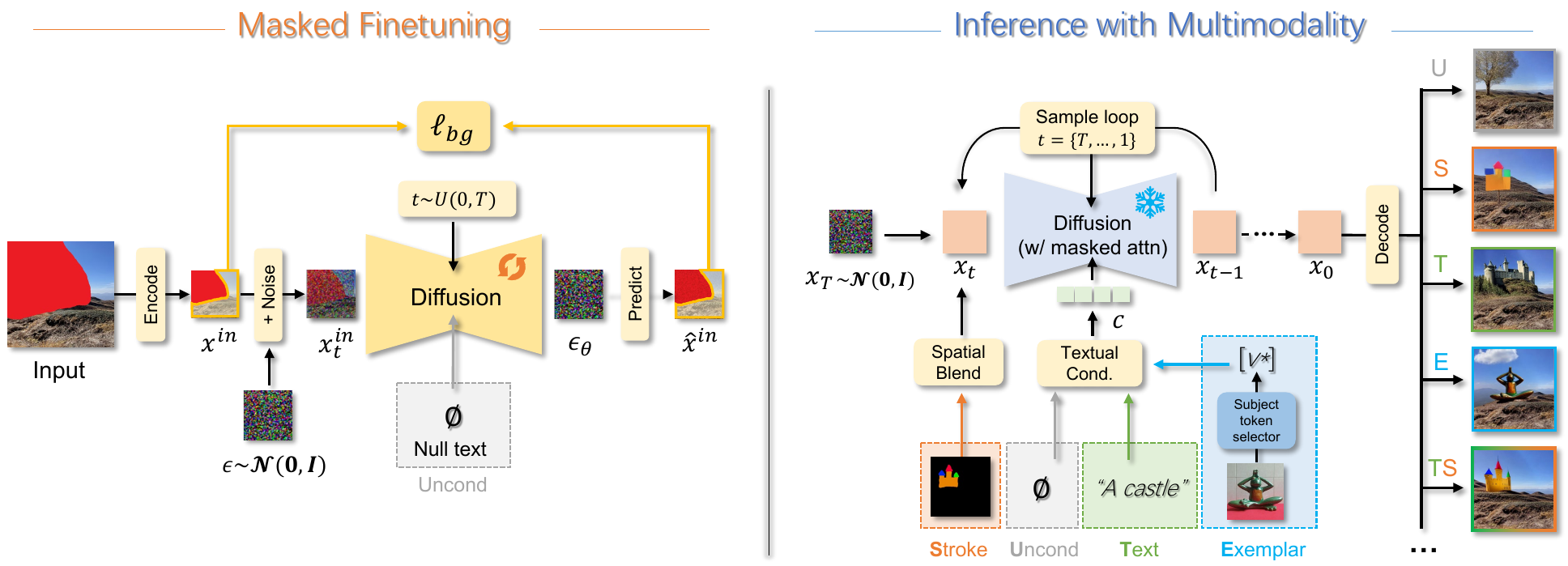}
\end{center}
\caption{Pipeline overview of Uni-paint. The model is finetuned on the known area of the input with null text. During sampling, unconditional, text-driven, stroke-driven and exemplar-driven inpainting can be achieved by conditioning on null text, text, stroke map, and exemplar's subject token, respectively.}
\label{fig.pipeline}
\end{figure*}

In our inpainting task, given an incomplete input image $x^{in}$ with a binary mask $m$ indicating its known region. Our uni-paint aims to inpaint its unknown part and outputs inpainted image $x^{out}$ unconditionally or conditioned on multimodal guidance, including text prompt $w$, exemplar image $x^{ref}$, stroke map $x^{stk}$, or even a mix of them, all based on a model parameterized by $\theta$.
This high-level process can be formulated as:
\begin{equation} \label{eq.unipaint}
x^{out} = \mathrm{Unipaint}_{\theta}(x^{in},m,[w,x^{ref},x^{stk}]),
\end{equation}
where $[\cdot]$ denotes optional conditional inputs. 
Technically, we implemented it through an iterative deterministic DDIM denoise sampling \cite{ddim} based on Stable Diffusion \cite{ldm}, where each denoise step $p(x_{t-1}|x_t,c)$ is formulated as:
\begin{equation} \label{eq.sample}
x_{t-1}=\sqrt{\alpha_{t-1}}\left(\frac{x_t-\sqrt{1-\alpha_t} \epsilon_\theta\left(x_t,c, t\right)}{\sqrt{\alpha_t}}\right)+\sqrt{1-\alpha_{t-1}} \cdot \epsilon_\theta\left(x_t,c, t\right),
\end{equation}
where $\alpha_t$ is time-dependent hyperparameter, $\epsilon_{\theta}$ is the noise predictor (i.e., the model) which takes current sampled image $x_t$, text embedding $c$, timestep $t$ as inputs, and predicts the noise for denoising sampling. The initial image $x_T$ is sampled from $\mathcal{N}(\mathbf{0,I})$ and the final denoised image $x_0$ will act as the output $x^{out}$.

\subsection{Overview} \label{sec.method.overview}

To leverage such model $\epsilon_{\theta}$ for inpainting, first we need to preserve the known region of $x^{out}$ as in $x^{in}$. A simple way is to blend the known part with $x_t$ during sampling as done in \cite{bld,bd}. However, we found this is insufficient since the known information is inserted externally rather than generated by the model itself, the model lacks full context awareness, potentially causing incoherent semantic transitions near hole boundary.
% another version: {
% the denoising process might not fully harmonize the stitching gap since the model lacks full context awareness, which can sometimes lead to incoherent semantic transitions near the hole boundary. 
% }
As such, we additionally finetune the model weights $\theta \to \theta^* $ such that it can inherently reconstruct the known region, as shown on left side of Fig.~\ref{fig.pipeline}. We demonstrate its superiority over direct blending in ablation study. 

The finetuned model $\epsilon_{\theta^*}$ is aware of the known region. To achieve multimodal inpainting in unknown region, according to Eq.~\ref{eq.sample}, we can alter the sampling by injecting semantic guidance via $c$, or spatial guidance via $x_t$, or not apply any guidance, depending on specific needs, as summarized on the right side of Fig.~\ref{fig.pipeline}. We inject the conditional information accordingly: 

\noindent(1) Unconditional inpainting can be achieved by leveraging model's own prior without any conditions ($c$ is set to null text $\emptyset$), formulated as $\epsilon_{\theta^*}(x_t,\emptyset,t)$, as illustrated in \textcolor{gray}{Uncond. block} in Fig.~\ref{fig.pipeline}.

\noindent(2) Given text $w$, we obtain its embedding $c$ through a text encoder $C$, the sampling is conditioned on $c$ via semantic interface, formulated as $\epsilon_{\theta^*}(x_t,C(w),t)$ as shown by \textcolor[RGB]{100,180,80}{Text block} in Fig.~\ref{fig.pipeline}.

\noindent(3) Given exemplar $x^{ref}$, despite being in image format, it is fundamentally different image composition. We expect the model to represent its distinct semantic features with some variation rather than naive copy-and-paste. Therefore, we associate $x^{ref}$ with an auto-selected token $v^*$ and inject it through semantic interface $c$, formulated as $\epsilon_{\theta^*}(x_t,C(v^*),t)$, as shown by \textcolor[RGB]{40,170,230}{Exemplar block}.

\noindent(4) Given stroke map $x^{stk}$, since it only requires color and spatial alignment while permitting ambiguous semantic interpretation, and text is struggled to deliver spatial information, so the best choice is to spatially inject $x^{stk}$ into $x_t$ to obtain modified $x'_t$ (see \textcolor[RGB]{240,130,70}{Stroke block}). The model operates as $\epsilon_{\theta^*}(x'_t,\emptyset,t)$.

These modalities can be used individually for single modality inpainting, or combined for multimodality inpainting, which we will introduce next.

%-------------------------------------------------------------------------
\subsection{Unconditional inpainting} \label{sec.method.uncond}
%-------------------------------------------------------------------------
We begin with unconditional inpainting as it serves as the foundation for conditional guidance.

%-----------
\noindent\textbf{Masked Finetuning}. 
Stable Diffusion has a strong generative prior learned from extensive pretraining, we leverage such prior for unconditional completion by introducing masked finetuning, where the model is finetuned to reconstruct the known part of the image. The masked finetuning enables the model to leverage its learned semantic understanding of the known areas, resulting in a plausible completion in unknown region.

Specifically, we adopt a typical training scheme as used in stable diffusion \cite{ldm}, which reduces the computational load by first having an encoder map the input image to low-dimension latent map (for the rest of this paper, notation $x$ refers to latent representation). The noising-denoising process is then performed in latent space, supervised by a simple noise loss derived in DDPM \cite{ddpm}. The difference is that we only calculate the loss on known region (specified by $m$). Our masked finetuning loss $\ell_{bg}$ is defined below:
\begin{equation}
\ell_{bg} = \mathbb{E}_{\epsilon,t}
\left\|m \odot \epsilon- m \odot \epsilon_\theta\left(x^{in}_t, \emptyset, t \right)\right\|_2^2,
\end{equation}
where $x^{in}_t$ is noised $x^{in}$ at timestep $t$: $x^{in}_t=\sqrt{\alpha_t}x^{in}+\sqrt{1-\alpha_t}\epsilon$ $(\epsilon \sim \mathcal{N}\mathbf{(0,I)})$, $\epsilon_\theta$ is the predicted noise from the model, $\emptyset$ is the null text embedding. Even if there is no explicit constraint on unknown area, the model is still able to  complete missing area in the sampling stage. This can be attributed to the natural coherence of the pre-training images, as well as the inherent inductive bias of convolution and self-attention layers to extend and replicate textures. 

%-----------
\noindent\textbf{Sampling with blending}. 

While finetuning the model helps memorize semantic content in known regions, yet does not guarantee perfect reconstruction, unless we overfit the known region through excessive finetuning, which can negatively impact the model's generative ability. To preserve the known region and the editability, we also employ blending technique,  where the known part of sampled latent $x_t$ is replaced by noised $x^{in}$ after each sampling step. This helps preserve the known region with much fewer iterations and less tuning on weights.

%-------------------------------------------------------------------------
\subsection{Text-driven inpainting} \label{sec.method.text}
%-------------------------------------------------------------------------
Once the model has been finetuned on known region, it will be immediately available for text-driven inpainting by conditioning on a text prompt $w$. Classifier-free guidance \cite{cfg} is also applied to boost the fidelity:
\begin{equation}
\hat{\epsilon}_{\theta^*}\left(x_t,c,t\right)=
\epsilon_{\theta^*}\left(x_t ,\emptyset,t \right) 
+s \left(\epsilon_{\theta^*}\left(x_t,c,t\right)-\epsilon_{\theta^*}\left(x_t , \emptyset, t \right)\right),
\end{equation}
where $c=C(w)$ is the embedding of $w$, $s$ is the guidance scale. Since we do not apply constraint on unknown region when finetuning, thus the model tends to respond to text roughly in the unknown area. Applying masked attention can bring more accurate control to editing scope, which will be discussed later.  
%-------------------------------------------------------------------------
\subsection{Exemplar-driven inpainting} \label{sec.method.exemplar}
%-------------------------------------------------------------------------
Exemplar-driven inpainting allows the user to provide an exemplar image $x^{ref}$ containing a subject, and the model should inpaint the subject in the unknown region while maintaining a reasonable semantic relationship with the background. Unlike image composition, the inserted subject should have the same identity as the exemplar but with variations rather than a simple copy-and-paste. Therefore, this is more challenging as it requires the model to remember exemplar's key semantic features.

%DreamBooth \cite{db} provides an insightful solution for exemplar-based generation by associating the exemplar with a subject token via finetuning. Similarly, 
To achieve this, we finetune the model on exemplar conditioned a class token $v^*$ that roughly corresponds to the subject. This process is performed in parallel with the unconditional finetuning by appending the following reference loss $\ell_{ref}$:
\begin{equation} 
\ell_{ref} = \mathbb{E}_{\epsilon,t}
\left\|\epsilon- \epsilon_\theta\left(x^{ref}_t, C(v^*), t \right)\right\|_2^2,
\end{equation}
where $x^{ref}_t$ is noised exemplar at timestep $t$, $C(v^*)$ returns the embedding of $v^*$. We do not include prior-preservation loss as done in DreamBooth \cite{db} since this can be unnecessary in our task, if users want to generate non-customized instances of a class, text-driven inpainting is a better alternative. To accelerate convergence and bring more diversity to object scale and position, we apply the augmentation by randomly scaling and shifting the exemplar image inside the unknown area's bounding box, and only calculate the loss on valid region. During the inference, conditioning on $C(v^*)$ allows the model to represent the subject.

%-----------
\noindent\textbf{Automatic subject identification}. 
 In normal penalization works \cite{db,txtinv}, users first need to manually specify an initial subject token that roughly describes the exemplar. Here we provide an alternative to automatically obtain the subject token $v^*$, which can be useful when users are unsure about the subject category or in automated scenarios. Since Stable Diffusion uses CLIP text model \cite{clip} as its text encoder, we use the corresponding CLIP image encoder to retrieve the token $v^*$ that best matches the exemplar. We represent this process as follows:
 \begin{equation} \label{eq.autosub}
v^* = \displaystyle\arg\max_{v_i \in \mathcal{V}}{\left(E_{T}(v_i) \cdot E_{I}(x^{ref})\right)},
 \end{equation}
where $\mathcal{V}$ is the set of all tokens, $E_{T}(v_i)$ is the CLIP text embedding of $v_i$, $E_{I}(x^{ref})$ is the  image embedding of $x^{ref}$. In practice, embedding set $\{E_{T}(v_i)\}$ can be pre-computed and stored, only a single inference of $E_{I}(x^{ref})$ is needed which costs negligible of time. We visualize such process in appendix Sec.~\ref{apx.autosub}.
 % where $\mathcal{V}$ is the set of all tokens, and $E_{T}(v_i)$ refers to the normalized embedding of token $v_i$ obtained from the CLIP text encoder, $E_{I}(x^{ref})$ is normalized embedding of $x^{ref}$ obtained from the CLIP image encoder. In practice, token embedding set $\{E_{T}(v_i)\}$ can be pre-computed, stored, and loaded as needed, only a single forward inference of $E_{I}(x^{ref})$ is needed which costs negligible of time.

%-------------------------------------------------------------------------
\subsection{Stroke-driven inpainting} \label{sec.method.stroke}
%-------------------------------------------------------------------------
Stroke-driven inpainting aims to generate real objects in unknown area that have a consistent shape and color with the user's stroke map. A finetuned model is also immediately available for stroke inpainting. To achieve this, we spatially blend the stroke latent $x^{stk}$ with sampled latent $x_t$ to obtain modified latent $x'_t$ at a certain intermediate timestep $\tau$ during the sampling, the blending operation $B$ is formulated as follows:
\begin{equation} \label{eq.stroke}
x'_t = B(x^{stk},x_t,t) = \\
\begin{cases}
    x_t\odot(1-m^{\mathit{stk}})  
     +  x^{stk}_t \odot m^{stk} &\text{if } t=\tau\\
    x_t  &\text{otherwise}
\end{cases}
\end{equation}
where $x^{stk}_t$ is noised stroke latent at time $t$, $m^{stk}$ is stroke mask derived from $x^{stk}$. It is noteworthy that our approach does not require users to scribble over the entire unknown area, but only the desired object, surrounding region will be completed automatically, benefiting from masked finetuning.

%-------------------------------------------------------------------------
\subsection{Multimodal inpainting} \label{sec.method.mixed}
%-------------------------------------------------------------------------
Our method supports inpainting with a mixture of the aforementioned guidance. Mixed semantic guidance (text + exemplar) can be achieved by simply conditioning on the embedding $c$ of combined exemplar token $v^*$ and text $w$, i.e., $c=C(w,v^*)$. Mixed semantic-spatial guidance can be performed by additional stroke blending. Specifically, we begin by conditioning on null text $\emptyset$ in the early stages ($t>\tau$), where null text helps with unconditional completion at these steps. When $t=\tau$, we perform spatial stroke blending as described in Eq.~\ref{eq.stroke}, followed by semantic conditioning $c$ for remaining steps ($t\leq\tau$). By adjusting $\tau$, we can control the trade-off between realism and stroke-faithfulness. We show additional results in ablation study.

% %
% While stroke-guided inpainting can be performed without text, we recommend incorporating text prompts to obtain more semantically meaningful results. To do so, we begin by conditioning on null text in the early stages ($t>\tau$), where null text helps with unconditional completion at these steps. When $t=\tau$, we perform the stroke blending as described in Eq.~\ref{eq.stroke}, followed by text conditioning during the remaining steps ($t\leq\tau$). By adjusting the time threshold $\tau$, we can control the trade-off between realism and stroke-faithfulness. We show additional results in ablation study.

%-------------------------------------------------------------------------
\subsection{Masked attention control} \label{sec.method.attnmask}
%-------------------------------------------------------------------------
While the model is finetuned to reconstruct the known region, we have observed an issue that the inpainted object may exceed the hole boundary and bleed into the known region, which may get truncated after blending, resulting in noticeable edge artifacts.

\begin{figure}[htbp]
\begin{center}

   \includegraphics[width=0.85\linewidth]{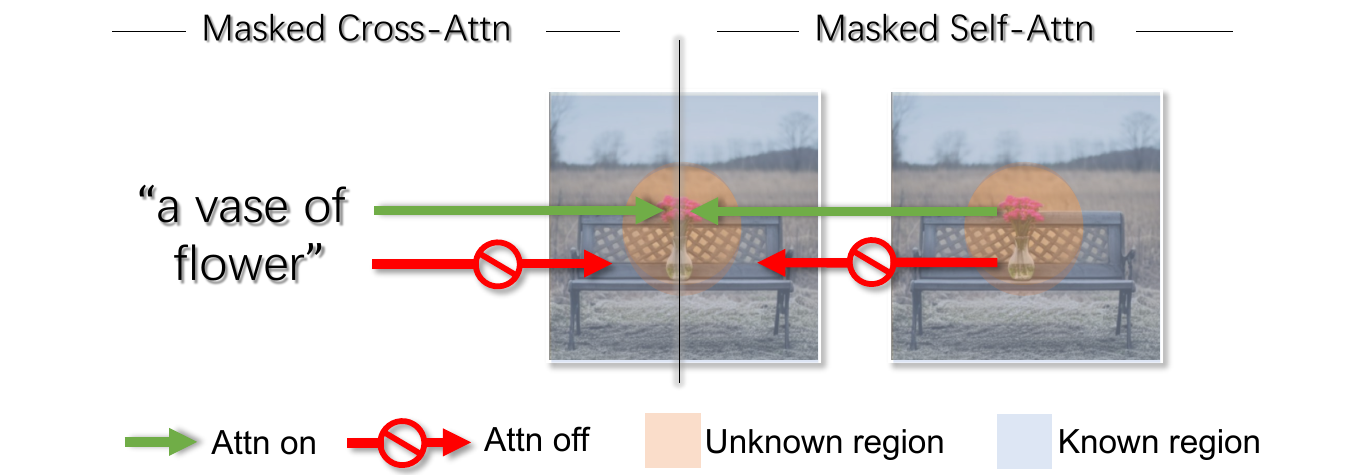}
\end{center}
   \caption{Illustration of masked attention control. For cross-attention (left), text \textcolor{green}{can} only attend to the unknown region but \textcolor{red}{not} the known region. For self-attention (right), inpainted content \textcolor{green}{can} only attend to the unknown region but \textcolor{red}{not} the known region.}
\label{fig.attn}
\end{figure}

To mitigate this issue, we introduce masked attention control, the general idea is to restrict text attention with the known region in cross-attention layers in the diffusion model, and restrict the inpainted region's attention with the known region for self-attention layers, as illustrated in Fig.~\ref{fig.attn}. Specifically, recall that in normal attention, query $Q$ , key $K$ and value $V$ are mapped from image features for self-attention, while for cross-attention, $K$, $V$ are mapped from textual features. Based on this, we introduce attention mask $M^{attn}$, and our masked attention is computed as follows:
\begin{equation}
\!\mathrm{MaskedAttn}(Q,K,V) = \left[\mathrm{softmax}\left(\frac{QK^T}{\sqrt{d}}\right) \odot M^{attn}\right] \times V \!
\end{equation}
where $d$ is the dimension of $K$ and $V$. Let $n$ be the image feature size and $l$ be the text feature length, and $\mathcal{I}$ be the set of known pixels' indices.

For cross-attention, $M^{attn}$ has the size of $n^2 \times l$. Here, $M^{attn}[i,j]$ denotes whether the $j^{th}$ textual token is allowed to attend to the $i^{th}$ image pixel. To prevent text from affecting known region, we set $M^{attn}[i,j]$ as follows:
\begin{equation}
M^{attn}[i,j] = 
\begin{cases}
0, & \text{if } i \in \mathcal{I} \\
1, & \text{if } i \notin \mathcal{I}
\end{cases}
\end{equation}

For self-attention, $M^{attn}$ has the size of $n^2 \times n^2$. $M^{attn}[i,j]$ indicates whether the $j^{th}$ image pixel is allowed to attend to the $i^{th}$ image pixel. To prevent inpainted pixels from leaking into known region, we set $M^{attn}[i,j]$ as follows:
\begin{equation}
M^{attn}[i,j] = 
\begin{cases}
0, & \text{if } j \notin \mathcal{I} \text{ and } i \in \mathcal{I} \\
1, & \text{otherwise}
\end{cases}
\end{equation}

We show in ablation studies that enabling masked attention control effectively prevents the generated object from overflowing outside the mask, thus avoiding truncation during blending.

%%%%%%%%%%%%%%%%%%%%%%%%%%%%%%%%%%%%%%%%%%%%%%%%%%%%%
\section{Experiments} \label{sec.exp}
%%%%%%%%%%%%%%%%%%%%%%%%%%%%%%%%%%%%%%%%%%%%%%%%%%%%%
%-------------------------------------------------------------------------
\subsection{Experiment setup} \label{sec.exp.impl}
%-------------------------------------------------------------------------
\noindent\textbf{Implementation details}. 
Our method is based on Stable Diffusion (sd-v1-4 checkpoint).
% \footnote{\url{https://github.com/CompVis/stable-diffusion}})
We finetuned the model for 100 iterations using Adam optimizer \cite{adam} with default parameters, and a learning rate of 1e-5. We use the DDIM sampler \cite{ddim} with 50 steps and a classifier-free scale of 8 \cite{cfg}.

\noindent\textbf{Datasets}. Our image samples were obtained from multiple sources, including: (1) EditBench \cite{editbench}, a systematic benchmark for text-driven image inpainting. We used its natural images and simple prompts as these are closer to practical use. (2) The \href{https://unsplash.com/}{Unsplash} website. (3) Images captured by ourselves. (4) Images from other papers.
%Our quantitative evaluation is based on (1) only whereas qualitative evaluation uses mixed sources.

\noindent\textbf{Evaluation metrics}.
We used the following popular quantitative metrics: (1) Neural Image Assessment (NIMA) \cite{nima}: a model-based reference-free perceptual image quality metric. (2) Text-to-image alignment (T2I) \cite{clipscore}: evaluates text-image CLIP similarity in text-driven inpainting. (3) Image-to-image alignment (I2I): evaluates the image-image CLIP similarity in exemplar-driven inpainting. (4) Root mean squared error (RMSE): assesses the faithfulness of the inpainted object to stroke color in stroke-driven inpainting. For these metrics, the test sample number varies from 50 to 120, depending on different tasks, and each sample's statistic is averaged from 8 diverse outputs. (5) Human preference: 55 participants were presented with 80 side-by-side result comparisons from different methods. To mitigate the choice bias, the order of the displayed options was randomized for each comparison. Participants were encouraged to vote for their most preferred result, but multiple selections were allowed (no more than half of the available options) if they found it hard to decide.

\noindent\textbf{Baselines}.
We compared our methods with state-of-the-art diffusion-based inpainting methods. For unconditional and text-driven inpainting, we compared with RePaint \cite{repaint} (unconditional only), SD-inpaint \cite{ldm}, GLIDE-inpaint \cite{glide}, and Blended Latent Diffusion (BLD) \cite{bld}. For stroke-driven inpainting, we compared with SDEdit \cite{sdedit}. For exemplar-driven inpainting, we compared with Textual Inversion (TxtInv) \cite{txtinv} and Paint-by-Example \cite{pbe}.

%-------------------------------------------------------------------------
\subsection{Text-driven/unconditional inpainting} \label{sec.exp.text}
%-------------------------------------------------------------------------

\begin{figure*}[htpb]
\begin{center}
   \includegraphics[width=\linewidth]{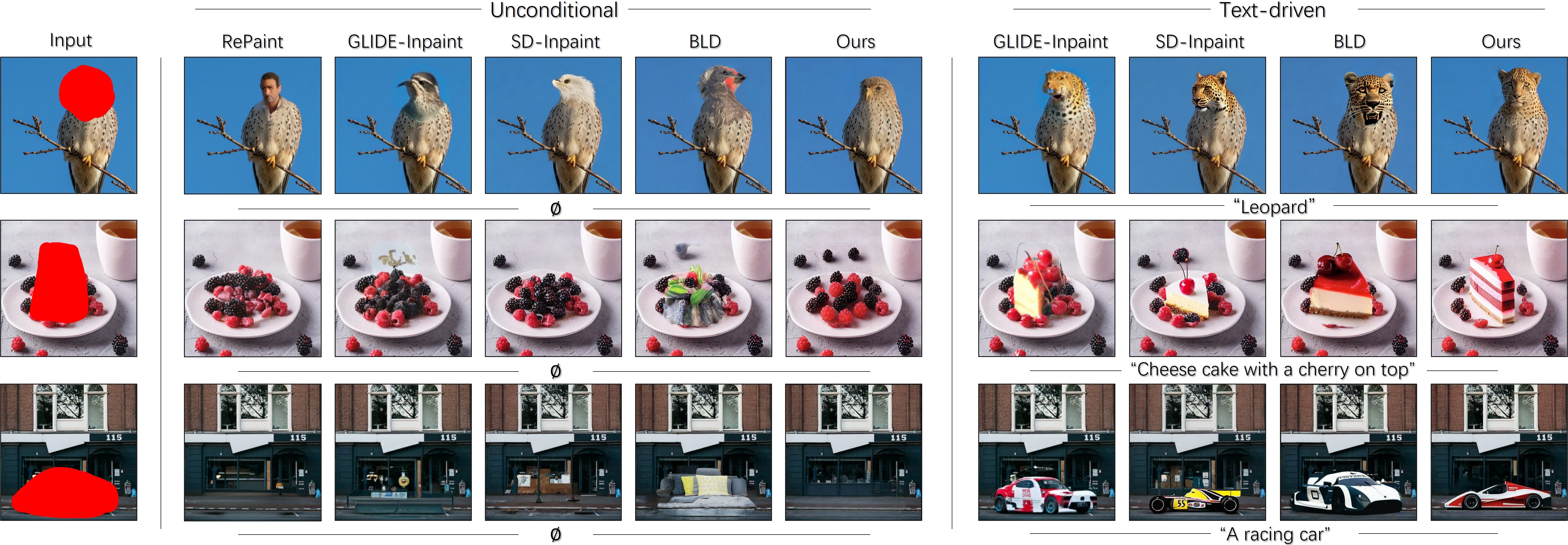}
\end{center}
   \caption{Qualitative comparison on unconditional (col.2-6) and text-driven (col.7-10) inpainting with related methods.}

\label{fig.text}
\end{figure*}

\noindent\textbf{Qualitative comparison}.%-----------------------
We present a side-by-side visual comparison with related baselines in Fig.~\ref{fig.text}. As can be seen, our method is able to generate more plausible results than GLIDE-inpaint, BLD, and RePaint. GLIDE-inpaint exhibits some artifacts along the hole boundary (as seen in the cheese cake example), while BLD tends to fit the mask shape, sometimes leading to unnatural transitions with the background (as seen with the over-sized leopard head). RePaint's repetitive sampling harmonizes the transitions but sometimes leads to incorrect inpainted semantics (see the bird with human head), since the image semantic is still not globally perceived by the model in essence. SD-inpaint and our method achieve comparable visual quality, but ours does not require training on a massive dataset. 
% We conjecture that since the released GLIDE model is trained on a reduced dataset for safety concerns, and is therefore not as powerful as Stable Diffusion. 

\noindent\textbf{Quantitative comparison}.%-----------------------
% During the evaluation, each method generated 8 diverse outputs for each input image-text pair in text-driven task. For unconditional task, the text was replaced with null text. 
We report T2I (which reflects faithfulness to text), NIMA (which focuses on technical quality but is agnostic to aesthetics), and human votes (which reflect individual subjective preference, as a complement to NIMA) of different methods in Tab.~\ref{tab.quant.text}. As can be seen, BLD obtains the highest T2I score but the lowest NIMA score, suggesting that it responds more strongly to the text, but sometimes this response is locally excessive and lead to unnatural oversized objects.
% This is less of a concern for the other tranin based methods, where the models possess full-context semantic awareness. Moreover, 
Our method shows comparable NIMA with other methods, but differs in the image-specific optimization, thus was favored by most human evaluators. 
Moreover, we noticed that SD-Inpaint shows lesser text-image alignment compared to SD base model (as its T2I score is lower than ours). This is because SD-Inpaint was trained with randomly generated masks, which often cover image region unrelated to the text prompt. Training on such masked images encourages the model to ignore the text, resulting in a reduced or even absent textual responses, especially when the masked regions are small. This finding is also revealed by recent studies \cite{smartbrush, editbench}. Therefore, our work builds on SD base model for its superior textual capability, which also brings more potential to other modalities.
 
\begin{table}[htbp]\small
\centering
\caption{Quantitative comparison on text-driven and unconditional (in parentheses) inpainting. }

\label{tab.quant.text}
\setlength{\tabcolsep}{2mm}
\begin{tabular}{@{}lccccc@{}}
\toprule
\textbf{}   & \multicolumn{1}{l}{T2I} & \multicolumn{1}{l}{NIMA}   &  \multicolumn{1}{l}{Human votes} \\ 
\midrule
% Ground truth          & 22.22      & 4.66 (4.66) & --  \\
% \midrule
RePaint \cite{repaint}       & -                & (4.48)  & (26.73\%) \\
BLD \cite{bld}               & \textbf{26.71}      & 4.25 (4.21) &  29.64\% (15.73\%)  \\
SD-Inpaint   \cite{ldm}      & 25.68       & \textbf{4.63} (\textbf{4.53}) &  32.82\% (36.73\%)\\
GLIDE-Inpaint  \cite{glide}  & 22.95        & 4.51 (4.45)  &  27.09\% (21.00\%)\\
Ours                         & 26.48      & 4.59 (4.49) &  \textbf{38.64}\%  (\textbf{56.82}\%)\\ 
\bottomrule
\end{tabular}
\end{table}

%-------------------------------------------------------------------------
\subsection{Exemplar-driven inpainting} \label{sec.exp.exemplar}
\begin{figure}[htpb]
\begin{center}
   \includegraphics[width=\linewidth]{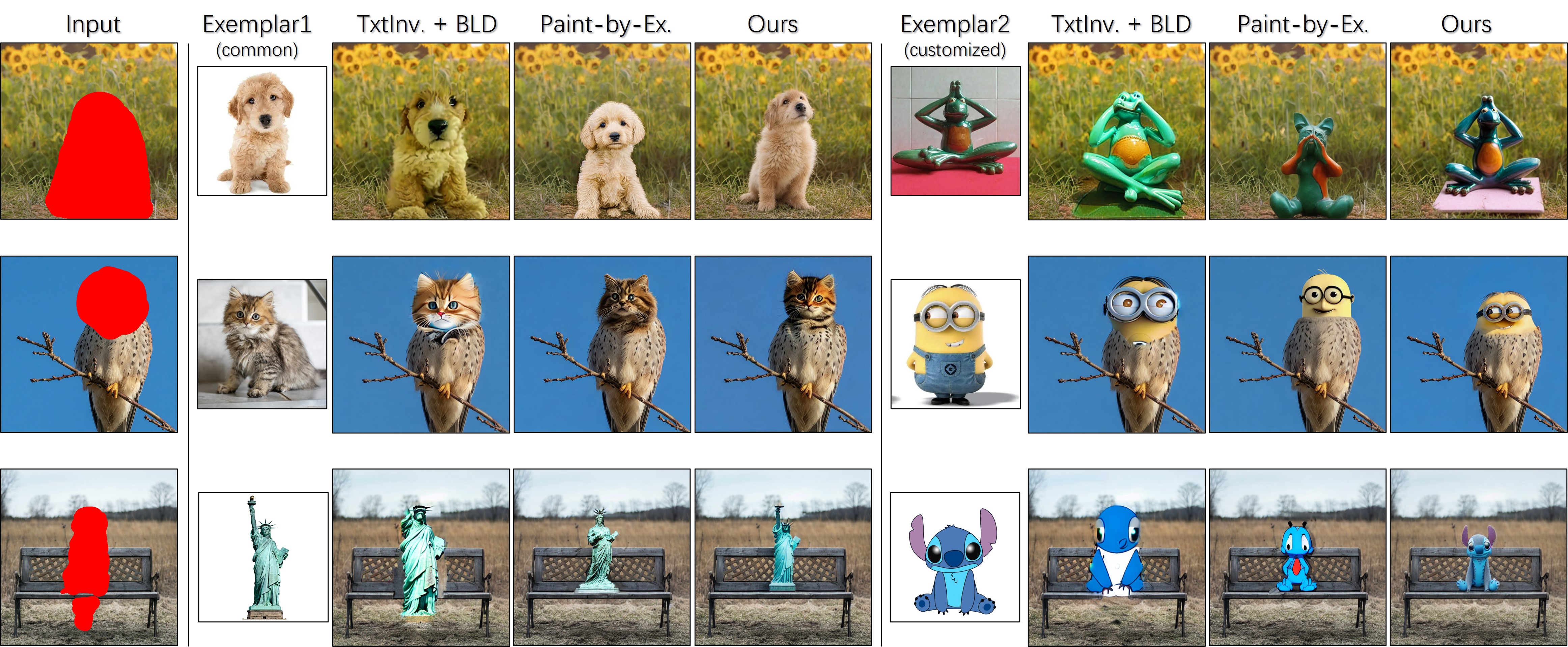}
\end{center}
   \caption{Qualitative comparison on exemplar-driven inpainting with related methods.} % Paint-by-Example \cite{pbe}.}
\label{fig.exemplar}
\end{figure}

%-------------------------------------------------------------------------
\noindent\textbf{Qualitative comparison}.  %-----------------------
We compared our results with Paint-by-Example \cite{pbe}, and the combined implementation of TxtInv and BLD as done in \cite{txtinv}, denoted as TxtInv+BLD. Our subject tokens were automatically determined using Eq.~\ref{eq.autosub}. The results are shown in Fig.~\ref{fig.exemplar}, where we can see our method can better retain details from the exemplar. Paint-by-Example achieves plausible results for commonly-seen concepts (e.g., cat and dog) but falls short when presented with less common or customized objects that may not appear in their training dataset. TxtInv+BLD inherits the similar truncation issues from BLD, and produces less aligned results than ours. We attribute this to the fact that fine-tuning the model normally has a stronger fitting ability compared to solely optimizing the word embedding.

% *** FIGURE \label{fig.exemplar} HAS BEEN MOVED TO ELSEWHERE, USE SEARCH TO FIND IT *****

\noindent\textbf{Quantitative comparison}.  %-----------------------
We use I2I score to measure the image similarly between the inpainted part and the exemplar image. We also report NIMA score and human preference in Tab.~\ref{tab.quant.exemplar}. While all these methods generate images of comparable quality, ours excels at capturing the semantics of the exemplar, particularly for personalized concepts, resulting in higher scores for both the I2I metric and human votes.

\begin{table}[htbp]\small
\centering
\caption{Quantitative results on exemplar-driven inpainting.}
\label{tab.quant.exemplar}
\setlength{\tabcolsep}{2.0mm}
\begin{tabular}{@{}lccccc@{}}
\toprule
\textbf{} & \multicolumn{1}{l}{I2I} & \multicolumn{1}{l}{NIMA} & \multicolumn{1}{l}{Human votes} \\ 
\midrule
% Exemplar image           & 100.00    &  4.79     & -- \\
% \midrule
TxtInv.+BLD \cite{txtinv,bld} & 78.24 & 5.25 & 9.36\% \\
Paint-by-Ex. \cite{pbe} & 77.75   & \textbf{5.32} & 21.27\% \\
Ours                    & \textbf{78.41 }   &5.28 & \textbf{69.36}\%\\
\bottomrule
\end{tabular}
\end{table}

%-------------------------------------------------------------------------
\subsection{Stroke-driven inpainting} \label{sec.exp.stroke}
%-------------------------------------------------------------------------

\begin{figure}[htpb]
\begin{center}
   \includegraphics[width=\linewidth]{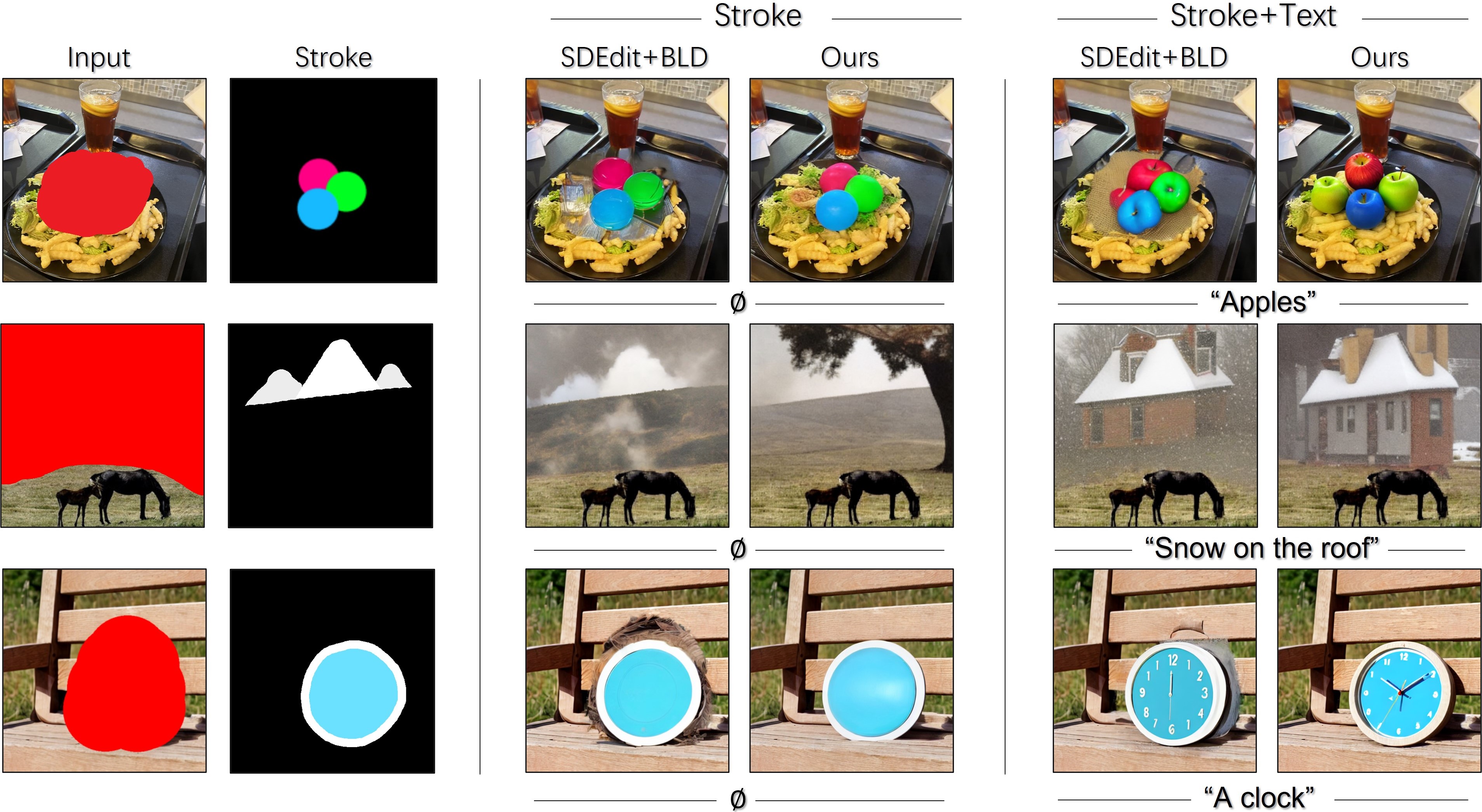}
\end{center}
   \caption{Qualitative comparison on stroke-driven inpainting.}
   % with the SDEdit+BLD baseline \cite{sdedit,bld}.}
\label{fig.stroke}
\end{figure}

\noindent\textbf{Qualitative comparison}. %-----------------------
Since there is no stroke-driven inpainting baseline so far, we compared our method with the combined implementation of SDEdit \cite{sdedit} and BLD \cite{bld}, denoted as SDEdit+BLD.
% SDEdit converts a stroke map into an image, while BLD preserves the known region. 
As shown in Fig.~\ref{fig.stroke}, SDEdit+BLD succeeds in generating objects that are well-aligned with the strokes. However, it fails to fill the remaining unknown area with plausible content where stroke hints are absent (see the apples in Fig.~\ref{fig.stroke}). In contrast, our approach accomplishes both stroke faithfulness and background completion, enabling users to focus on their interested objects without having to scribble over the entire missing area. This reveals that the proposed masked finetuning helps the model gain awareness of image semantics, bringing plausible completion in the unknown region.

\begin{figure}[htpb]
\begin{center}
   \includegraphics[width=\linewidth]{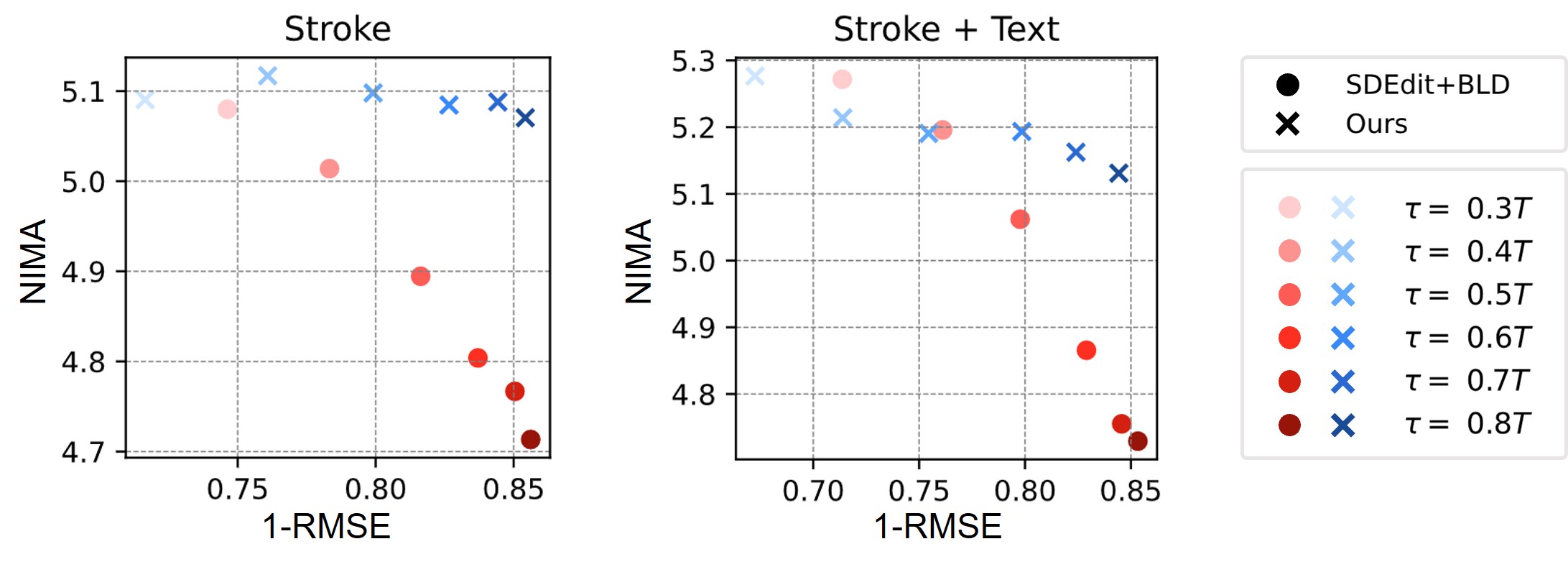}
\end{center}
   \caption{Quantitative comparison of inpainting quality (NIMA score) and stroke-alignment (1-RMSE) on stroke-driven (left) and stroke+text-driven (right) inpainting with different $\tau$. Upper right corner indicates a better trade-off.}
\label{fig.stroke.quanti}
\end{figure}

\begin{figure*}[htpb]
\begin{center}
   \includegraphics[width=\linewidth]{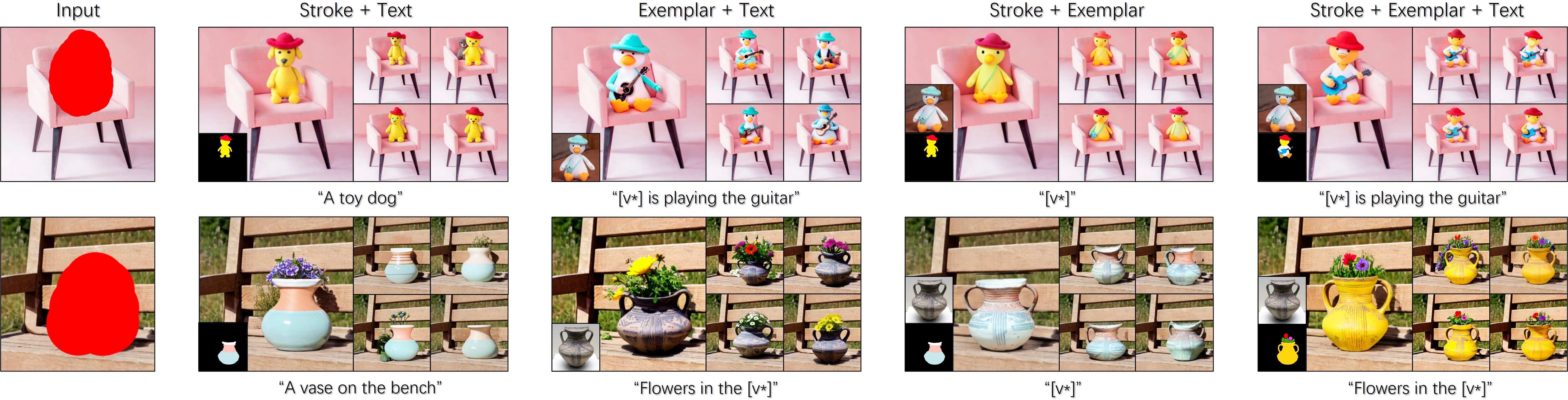}
\end{center}
   \caption{Diverse inpainting results under mixed guidance from a combination of text, stroke, and exemplar.}
\label{fig.mixed}
\end{figure*}

\noindent\textbf{Quantitative comparison}.  %-----------------------
Fig.~\ref{fig.stroke.quanti} presents quantitative statistics on two aspects: color faithfulness to stroke, measured by 1-RMSE (flipped RMSE), and image quality, measured by NIMA score. There exits a trade-off between them when choosing different stroke blending timestep $\tau$. As can be seen, at the same level of stroke faithfulness, our results generally have a better quality. We attribute this to better background preservation and higher degree of editing in our approach. For human evaluation, our method received 54.64\% of the votes as opposed to 45.36\% of SDEdit+BLD.

% \begin{figure}[htpb]
% \begin{center}
%    \includegraphics[width=\linewidth]{fig.stroke.quanti.jpg}
% \end{center}
%    \caption{Quantitative comparison of inpainting quality (NIMA score) and stroke-alignment (1-RMSE) on stroke-driven (left) and stroke+text-driven (right) inpainting with different $\tau$. Upper right corner indicates a better trade-off.}
% \label{fig.stroke.quanti}
% \end{figure}

%-------------------------------------------------------------------------
\subsection{Inpainting with mixed guidance} \label{sec.exp.mixed}
%-------------------------------------------------------------------------
Uni-paint stands out from previous approaches by supporting the use of mixed multimodal guidance for inpainting task. We demonstrate this capability in Fig.~\ref{fig.mixed}, where text or exemplar are used to deliver the subject's semantic attributes and stroke is used to determine its color and layout. By sampling from different initial noise, our method can generate diverse outputs given the same input and guidance. %We observe that the presence of stroke guidance can affect the inpainting diversity, as the spatial layout is constrained.

% *** FIGURE \label{fig.mixed} HAS BEEN MOVED TO ELSEWHERE, USE SEARCH TO FIND IT *****

%-------------------------------------------------------------------------
\subsection{Ablation studies} \label{sec.exp.ablation}
%-------------------------------------------------------------------------
We conduct ablation studies on several settings used in our work.

\begin{figure}[htpb]
\begin{center}
   \includegraphics[width=\linewidth]{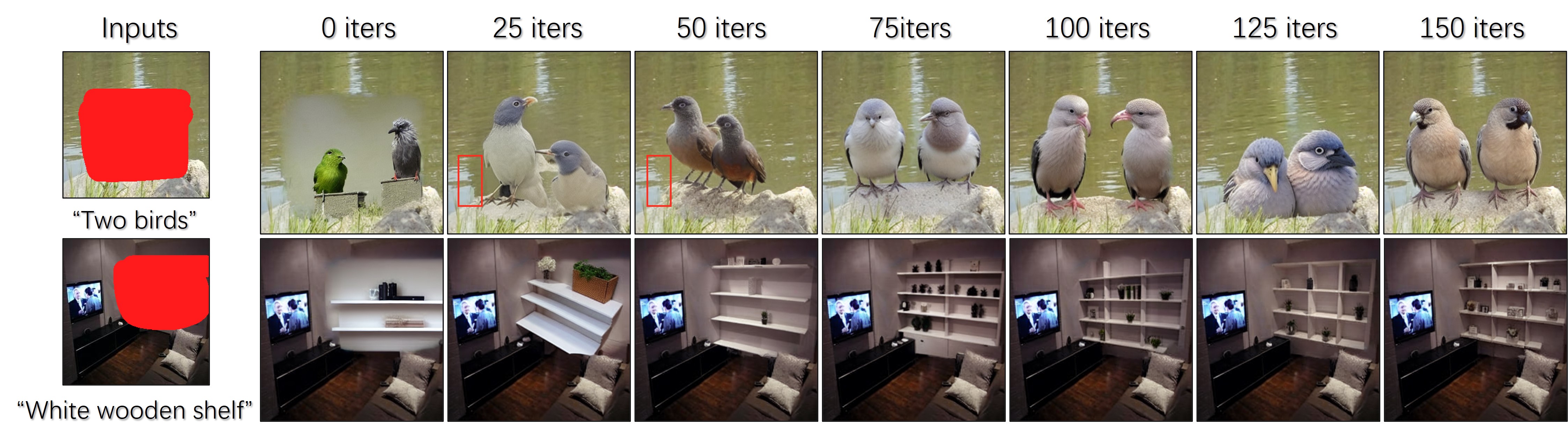}
\end{center}
   \caption{Effect of masked finetuning with different iterations. Applying masked finetuning enhances model's context awareness and brings better texture transition.}
\label{fig.abl.finetune}
\end{figure}

%-----------  %------------------------
\noindent\textbf{Masked finetuning}. 
To demonstrate the benefits of masked finetuning, we present visual examples with different finetuning iterations in Fig.~\ref{fig.abl.finetune}. Without masked finetuning (0 iters), the model only focuses on the local region and disregards the context, resulting in noticeable stitching artifacts. This issue is mitigated after 75 iterations of finetuning. This suggests that masked finetuning helps the model gain semantic awareness in the known region, leading to coherent texture transitions in inpainted region.

\begin{figure}[htpb]
\begin{center}
   \includegraphics[width=\linewidth]{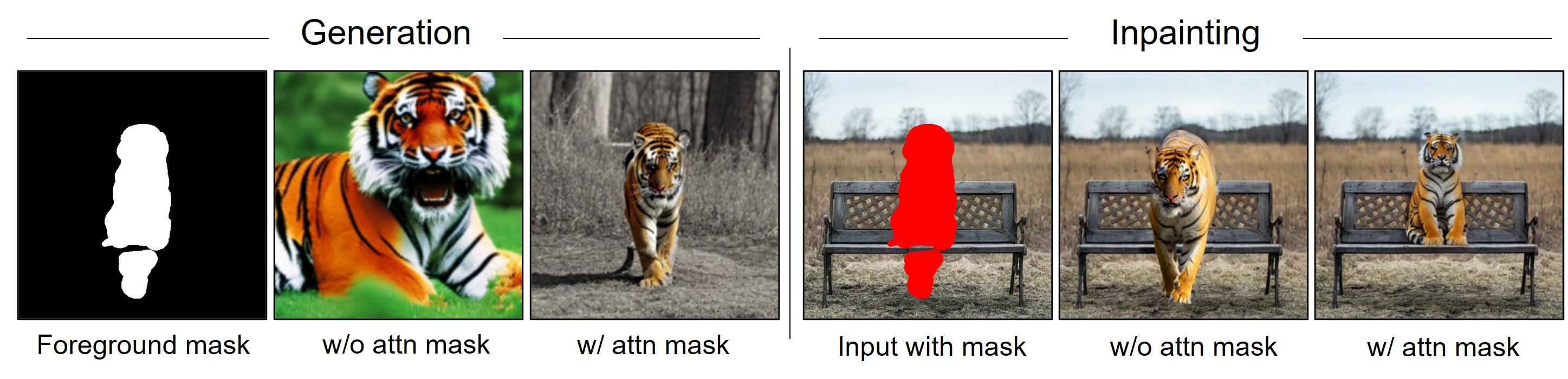}
\end{center}
   \caption{Examples of generation/inpainting of tiger with/without masked attention control. Applying masked attention control effectively constrains generation scope.}
\label{fig.abl.attn}
\end{figure}

%-----------  %------------------------
\noindent\textbf{Masked attention control}. 
We introduce masked attention control mainly to suppress over-sized inpainted content from leaking into the known area. To demonstrate this effect, we generate images of tigers with and without applying masked attention in two scenarios: free-generation and inpainting, as shown in Fig.~\ref{fig.abl.attn}. In the free-generation case, enabling masked attention control effectively constrains the generation scope within the masked region. In the inpainting case, disabling masked attention control can sometimes result in over-sized inpainted content, which may get truncated after background blending. This suggests that restricting the attention flow can be useful in local editing tasks like inpainting.

\begin{figure}[htpb]
\begin{center}
   \includegraphics[width=\linewidth]{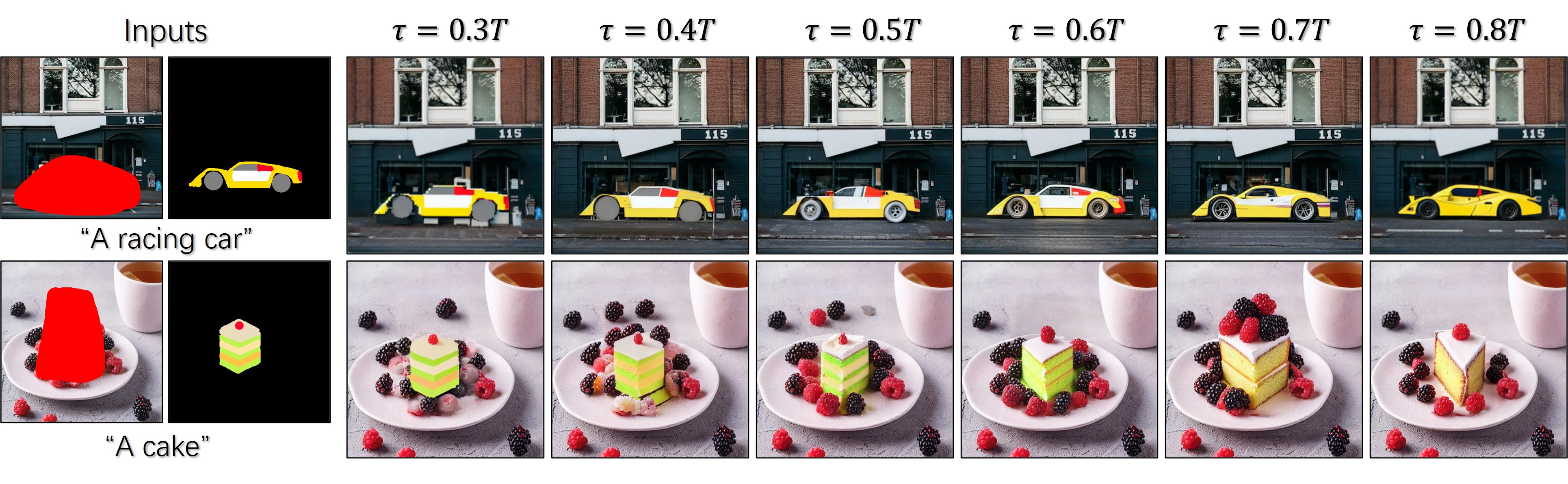}
\end{center}
   \caption{Effect of different $\tau$ on stroke-driven inpainting. Larger $\tau$ leads to more realistic but less aligned results.}
\label{fig.abl.stroke}
\end{figure}

%-----------  %------------------------
\noindent\textbf{Stroke blending timestep}. 
In stroke-driven inpainting, threshold $\tau$ adjusts the balance between realism and stroke-faithfulness, we show a series of visual results with different choices of $\tau$ in Fig.~\ref{fig.abl.stroke}, quantitative statistics can be found in Fig.~\ref{fig.stroke.quanti}. Generally, larger $\tau$ leads to more realistic but less aligned results, $\tau \in [0.5T,0.6T]$ normally yields a balanced effect ($T$ is the total number of timestep).

\begin{figure}[htpb]
\begin{center}
   \includegraphics[width=1.0\linewidth]{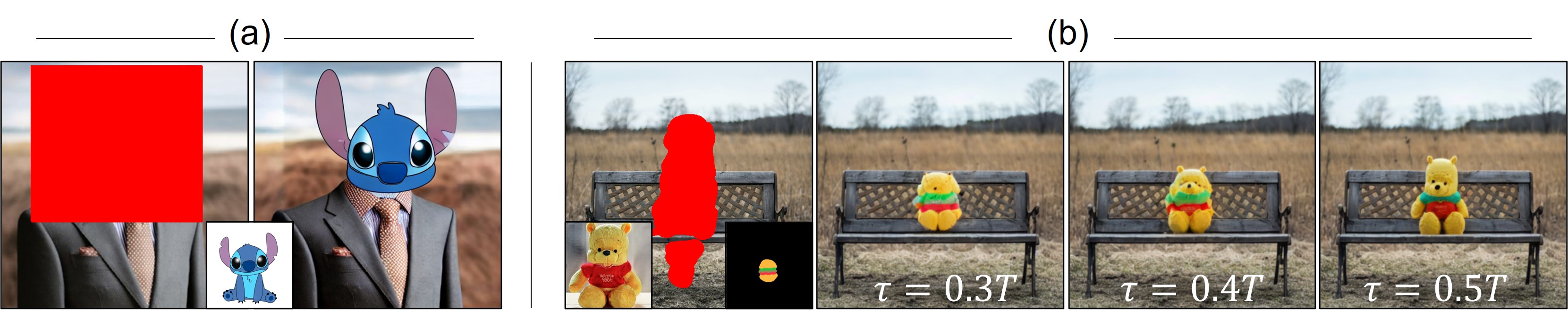}
\end{center}
   \caption{Failure cases: (a) Unnatural stitching. (b) Failed to obey conflicting guidance (stroke shape and exemplar)}%, where the stroke shape (burger) significantly deviates from the exemplar (bear).}
\label{fig.limitaion}
\end{figure}

%%%%%%%%%%%%%%%%%%%%%%%%%%%%%%%%%%%%%%%%%%%%%%%%%%%%%
\section{Conclusion} \label{sec.conclusion}
%%%%%%%%%%%%%%%%%%%%%%%%%%%%%%%%%%%%%%%%%%%%%%%%%%%%%

We propose Uni-paint, a unified multimodal inpainting framework supporting unconditional, text, stroke, and exemplar guidance. Our unconditional and text-driven inpainting results are competitive with recent works without large-scale training. For exemplar-driven inpainting, our few-shot approach achieves improved customization effects. Stroke guidance on regions of interest is also integrated in our framework. Moreover, our method supports inpainting with mixed guidance, which is not available in existing methods.

However, our method still encounters some limitations. First, when there is a large gap between the exemplar and the input (e.g., cartoon vs. real images), our method may fail to fully harmonize the gap, resulting in unnatural stitching (see Fig.\ref{fig.limitaion}a). Second, conflicts may occur when mixing guidance from different modalities. For example, in Fig.\ref{fig.limitaion}b, the hamburger-shaped stroke is far different from the exemplar, making it challenging to find an appropriate $\tau$ that simultaneously respects both the stroke and the exemplar guidance. Note this issue can be avoided with careful interactions.
% Note that these two problems can be avoided with more careful user interactions.

Our future work aims to address these limitations and explore additional modalities, further investigating the potential of diffusion models in image inpainting.

%%
%  ** Social Impact is not required by ACM MM offical 
%%


\begin{acks}
This work is supported by GRF grant %(Project No. CityU 11208123) 
from the Research Grants Council (RGC) of Hong Kong. We also thank Unsplash and the photographers for generously sharing their high-quality, free-to-use images used in this research.
\end{acks}

\bibliographystyle{ACM-Reference-Format}
\balance
\bibliography{mybib}

%%%%%%%%%%%%%%%%%%%%%%%%%%%%%%%%%%%%%%%%%%%%%%%%%%%%%%%%%%%%%%%%%%%%%%%%%%%%%%%%%%%%%%%%%%%%%%%%%%%%%%%%%%%%%%%%%%%%%%%%%%%%%%%%%%%%%%%%%%%%%%%%%%%%%%%%%%%%%%%%%%%%%%%%%%%%%%%%%%%%%%%%%%%%%%%%%%%%%%%%%%%%%%%%%%%%%%%%%%%%%%%%%%%%%%%%%%%%%%%%%%%%%%%%%%%%%%%%%%%%%%%%%%%%%%%%%%%%%%%%%%%%%%%%%%%%%%%%%%%%%%%%%%%%%%%%%%%%%%%%%%%%%%%%%%%%%%%%%%%%%%%%%%%%%%%%%%%%%%%%%%%%%%%%%%%%%%%%%%%%%%%%%%%%%%%%%%%%%%%%%%%%%%%%%%%%%%%%%%%%%%%%%%%%%%%%%%%%%%%%%%%%%%%%%%%%%%%%%%%%%%%%%%%%%%%%%%%%%%%%%%%%%%%%%%%%%%%%%%%%%%%%%%%%%%%%%%%%%%%%%%%%%%%%%%%%%%%%%%%%%%%%%%%%%%%%%%%%%%%%%%%%%%%%%%%%%%%%%%%%%%%
\clearpage
\appendix

\noindent{\Huge{\textbf{Appendix}}} \\

In this appendix, we provide additional details and results of our approach. We present more implementation details in Sec.~\ref{apx.impl}. Additional human evaluation details are given in Sec.\ref{apx.userstudy}. We also provide more visual result of unconditional, text-driven, stroke-driven, exemplar-driven, and multimodality inpainting in Sec.~\ref{apx.vis}.% Furthermore, we make an interactive demo for our approach, as shown in Sec.~\ref{apx.demo}.

%%%%%%%%%%%%%%%%%%%%%%%%%%%%%%%%%%%%%%%%%%%%%%%%%%%%%%%%%%%
\section{Implementation details} \label{apx.impl}

\subsection{Masked finetuning}  \label{apx.impl.finetune}
We introduced our general motivation and core steps of masked finetuning in Sec. 3.2 of the main paper. Here we present more detailed process in pseudo code format in Algorithm \ref{alg.finetune}.

\begin{algorithm}
\caption{Masked finetuning} \label{alg.finetune}
\begin{algorithmic}[1]
\Require
% \Statex \textbf{Require}
Input image $X^{in}$, binary mask $M$, exemplar $X^{ref}$ (optional), 
pretrained stable diffusion model $\epsilon_{\theta}$, text encoder $C$, image encoder $E$.

\State Get image latent $x^{in} = E(X^{in}\odot M)$
\State Get latent mask $m=\mathrm{Resize}(M)$ s.t. $m$ has the same size as $x^{in}$
\State Get null text embedding $\emptyset=C(\text{""})$
\State Get exemplar token $v^*$ from Eq. 6 if  $X^{ref}$ exists
%\Procedure{ExampleProcedure}{}
\While{$iter< total\_iters$}
\State $t_1 \sim \mathcal{U}(0,T) $
\State $\epsilon_1 \sim \mathcal{N}(\textbf{0,I}) $
\State $x^{in}_{t_1}=\sqrt{\alpha_{t_1}}x^{in}+\sqrt{1-\alpha_{t_1}}\epsilon_1$
\State $\ell_{bg} = \left\|m \odot \epsilon_1- m \odot \epsilon_\theta\left(x^{in}_{t_1}, \emptyset, t_1 \right)\right\|_2^2 $
\If{exist($X^{ref}$)}
\State $t_2 \sim \mathcal{U}(0,T) $
\State  $X^{ref}, M^{ref} = \mathrm{RandomShiftAndScale}(X^{ref})$  %\/\/ Exemplar augmentation
\State $x^{ref} = E(X^{ref})$  % Get exemplar latent 
\State $m^{ref} = \mathrm{Resize}(M^{ref})$
\State $\epsilon_2 \sim \mathcal{N}(\textbf{0,I}) $
\State $x^{ref}_{t_2}=\sqrt{\alpha_{t_2}}x^{ref}+\sqrt{1-\alpha_{t_2}}\epsilon_2$
\State $\ell_{ref} = \left\|m^{ref} \odot \epsilon_2- m^{ref} \odot \epsilon_\theta\left(x^{ref}_{t_2}, C(v^*), t_2 \right)\right\|_2^2 $
\Else
\State $\ell_{ref} =0$
\EndIf
\State $\ell = \ell_{bg} + \ell_{ref}$
\State $\theta = \theta - lr \cdot \nabla_{\theta}\ell$
\EndWhile
\Statex \textbf{Return} Finetuned model $\epsilon_{\theta^*}$
\end{algorithmic}
\end{algorithm}

\begin{figure*}[htpb]
\begin{center}
   \includegraphics[width=1.0\linewidth]{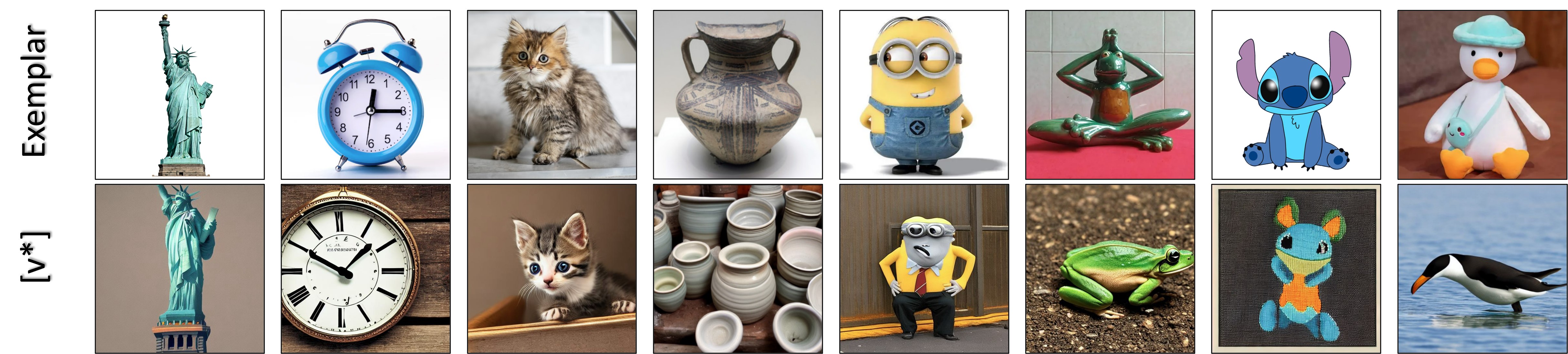}
\end{center}
   \caption{Visualization examples of automatic subject token identification. The bottom-row images are generated by \textbf{un-tuned} model conditioned on subject token $v^*$ obtained from the top-row exemplar images, suggesting that $v^*$ is able to provide a rough initial approximation to the exemplar. }
\label{fig.apx.autosub}
\end{figure*}

%%%%%%%%%%%%%%%%%%%%%%%%%%%%%%%%%%%%%%%%%%%%%%%%%%%%%%%%%%%
\subsection{Auto-subject token visualization} \label{apx.autosub}
In Sec. 3.4, we developed a CLIP-based method for automatically obtaining the initial subject token $v^*$ from a given exemplar, which can be useful in cases where the user is uncertain about the subject category or in automated application scenarios. Note that before finetuning the model, $v^*$ only provides a rough initial approximation but not perfect alignment to the exemplar. We present some visualization examples of this approach in Fig.~\ref{fig.apx.autosub}. In these examples, the bottom-row images are generated conditioned on $v^*$ by an \textbf{un-tuned} model obtained from the top-row exemplar images. We found that for well-known concepts such as the Statue of Liberty, $v^*$ is directly capable of reproducing the exemplar. For less-common concepts (e.g., cartoon characters Stitch and Minion), while the details are not perfectly preserved, $v^*$ still provides a rough initial approximation for capturing the exemplar concepts. After the model is fine-tuned, $v^*$ will be bound with exemplar and is able to produce aligned results.

\subsection{Computational speed}
With a batch size of 1, our model takes roughly 98 seconds for finetuning with 100 iterations, and 4.8 seconds for inference with 50 DDIM steps on a NVIDIA A6000 GPU, which is the typical speed of official released stable diffusion model without using acceleration strategy or model compression. Note that the speed can be further accelerated by using more advanced sampler (e.g., DPM-Solver++) or toolbox (e.g., xFormers).

\section{Human evaluation details} \label{apx.userstudy}
\subsection{Setup}
\begin{figure}[htpb]
\begin{center}
   \includegraphics[width=1.0\linewidth]{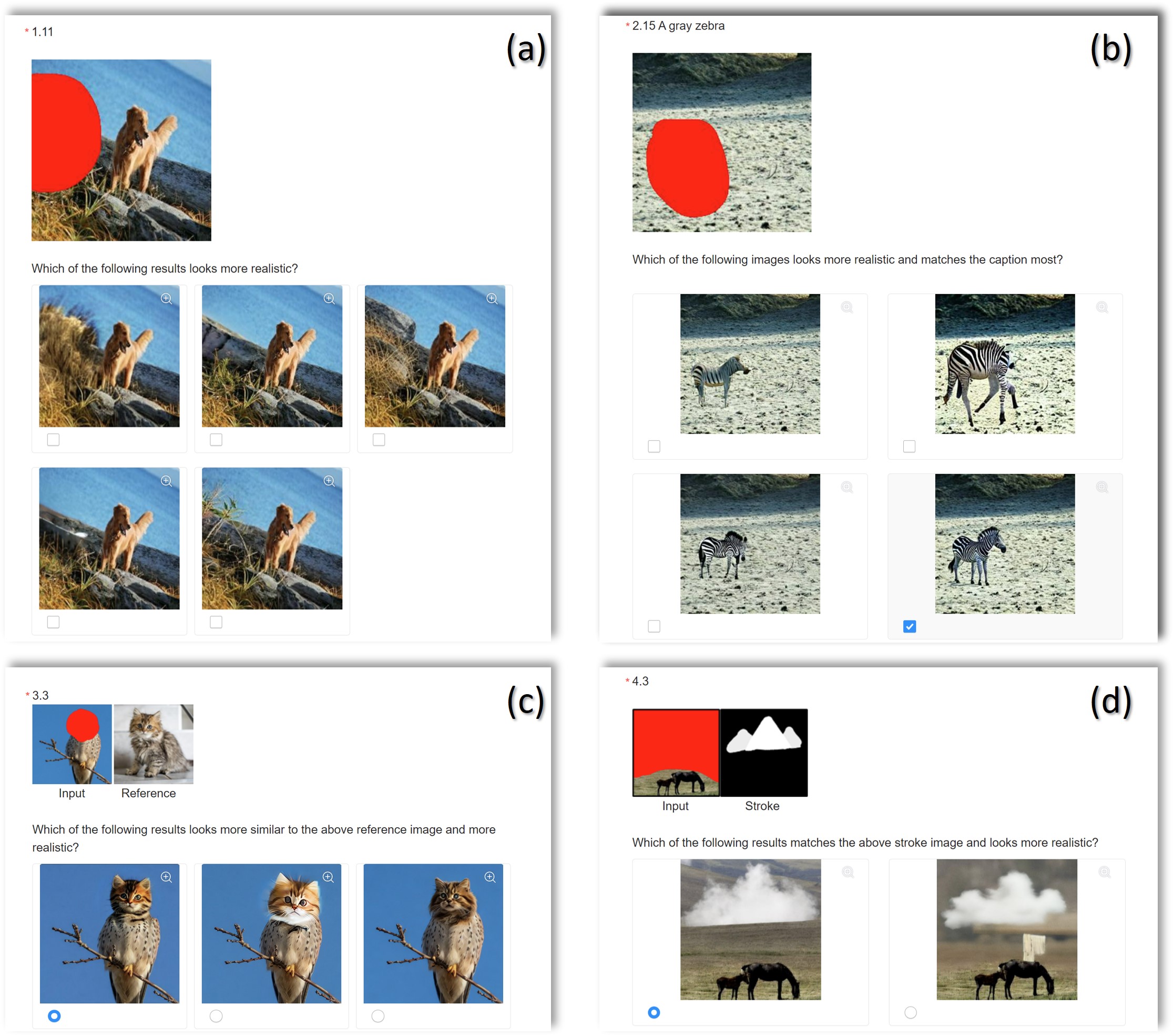}
\end{center}
   \caption{Interface of human evaluation questionnaire, these are four example questions of four different inpainting task: (a) Unconditional inpainting, (b) Text-driven inpainting, (c) Exemplar-driven inpainting, (d) Stroke-driven inpainting. Participants can select up to 2 options for (a) and (b), and 1 option only for (c) and (d).}
\label{fig.apx.userstudy.ui}
\end{figure}

As described in Sec. 4.1 in the main paper, we conducted a user study in questionnaire format to determine which method produces the best results in terms of human perception. We invited a total of 55 participants. None of the participants were involved in this research in part or in whole, or had any conflicts of interest. The questionnaire consists of 4 sections for 4 different guidance: unconditional inpainting, text-driven inpainting, exemplar-driven inpainting, and stroke-driven inpainting, respectively. Each section has 20 side-by-side comparisons of different methods. To mitigate the potential choice bias, the displayed order of options was randomly shuffled for each question and each participant. For each question, participants were encouraged to choose their most preferred option, but in case they found it hard to decide, they were allowed to make multiple choices but no more than half of the available options (i.e., up to 2 options for unconditional and text-driven tasks, and 1 option only for exemplar-driven and stroke-driven tasks). Fig.~\ref{fig.apx.userstudy.ui} shows some example questions from our questionnaire. 

\subsection{Statistics}
\begin{figure*}[htpb]
\begin{center}
   \includegraphics[width=\linewidth]{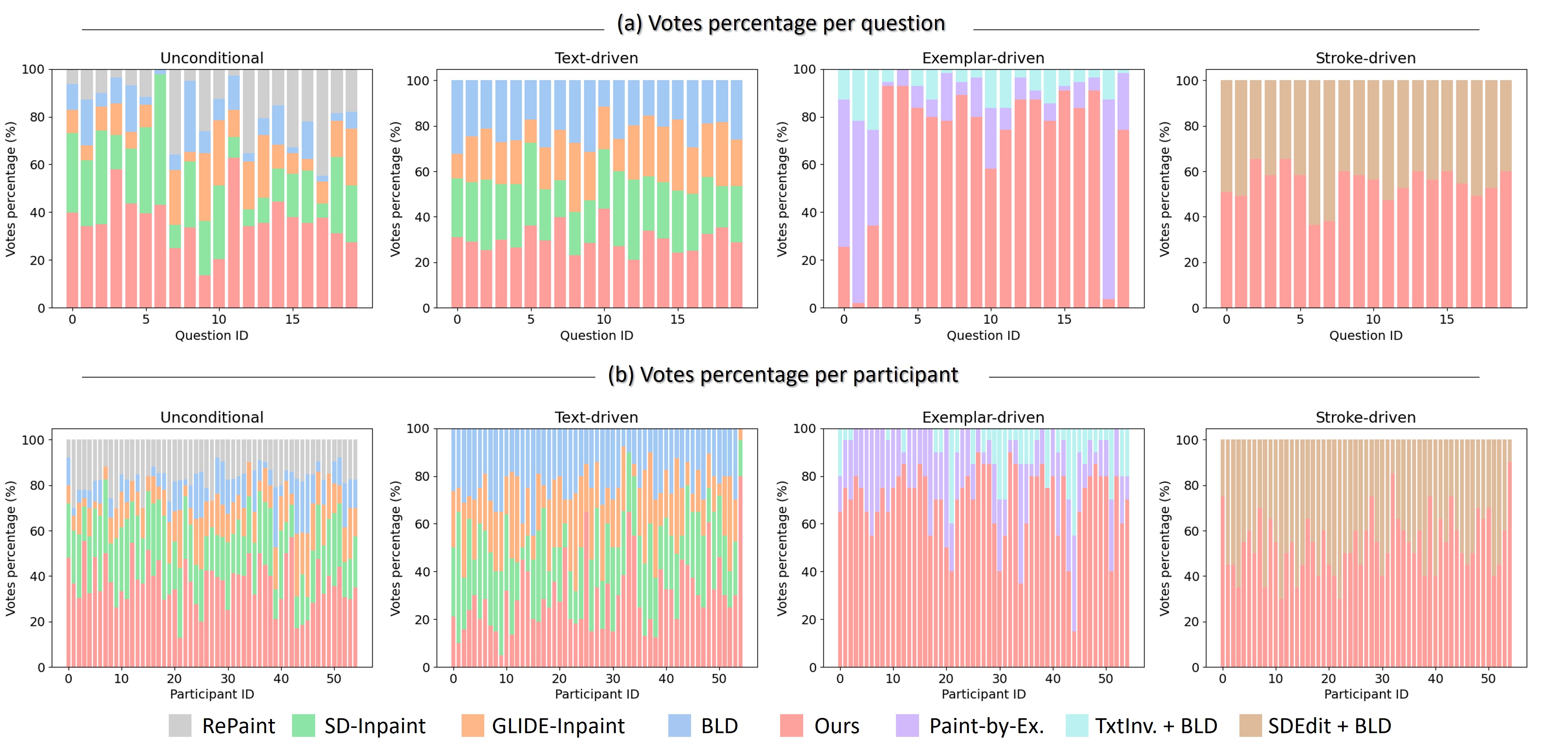}
\end{center}
   \caption{Detailed human evaluation statistics. (a) presents the votes percentage of each question. (b) presents the votes percentage of each participant.}
\label{fig.apx.userstudy.quanti}
\end{figure*}

We also show detailed human evaluation statistics from two aspects: votes percentage per question (refers to the percentage of votes received by a method, relative to the total number of votes for a question), and votes percentage per participant (refers to the percentage of votes received by a method, relative to the total number of votes cast by a participant), as demonstrated in Fig.~\ref{fig.apx.userstudy.quanti}. As can be seen, our unconditional and text-driven inpainting results are roughly comparable to SD-Inpaint, with a slight numerical advantage. Our exemplar-driven and stroke-driven results received more votes in a greater number of questions and were favored by more of voters.

%%%%%%%%%%%%%%%%%%%%%%%%%%%%%%%%%%%%%%%%%%%%%%%%%%%%%%%%%%%
\section{Additional visual results}\label{apx.vis}
\begin{figure*}[htpb]
\begin{center}
   \includegraphics[width=\linewidth]{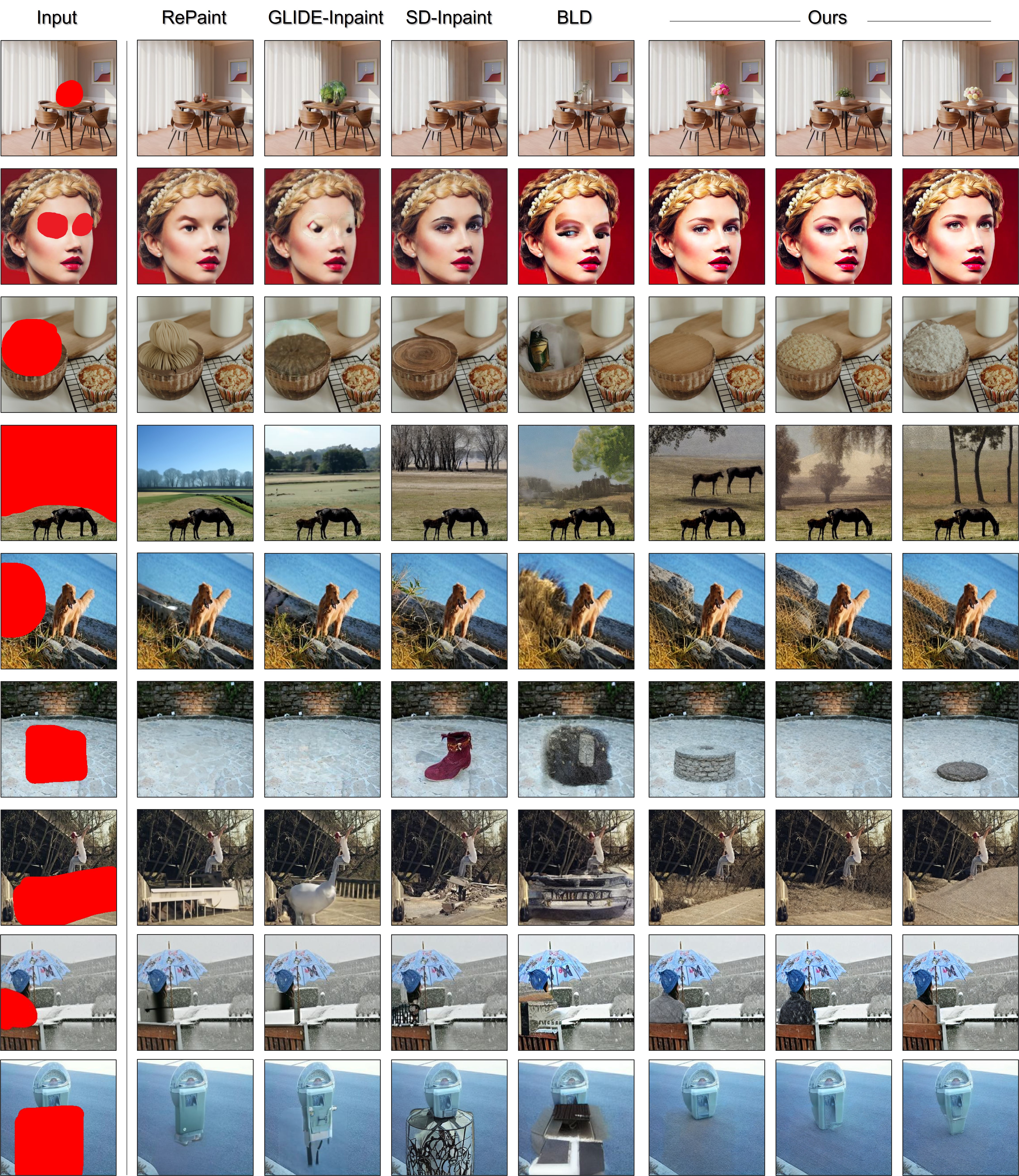}
\end{center}
   \caption{Additional unconditional inpainting results.}
\label{fig.apx.uncond}
\end{figure*}

\begin{figure*}[htpb]
\begin{center}
   \includegraphics[width=0.9\linewidth]{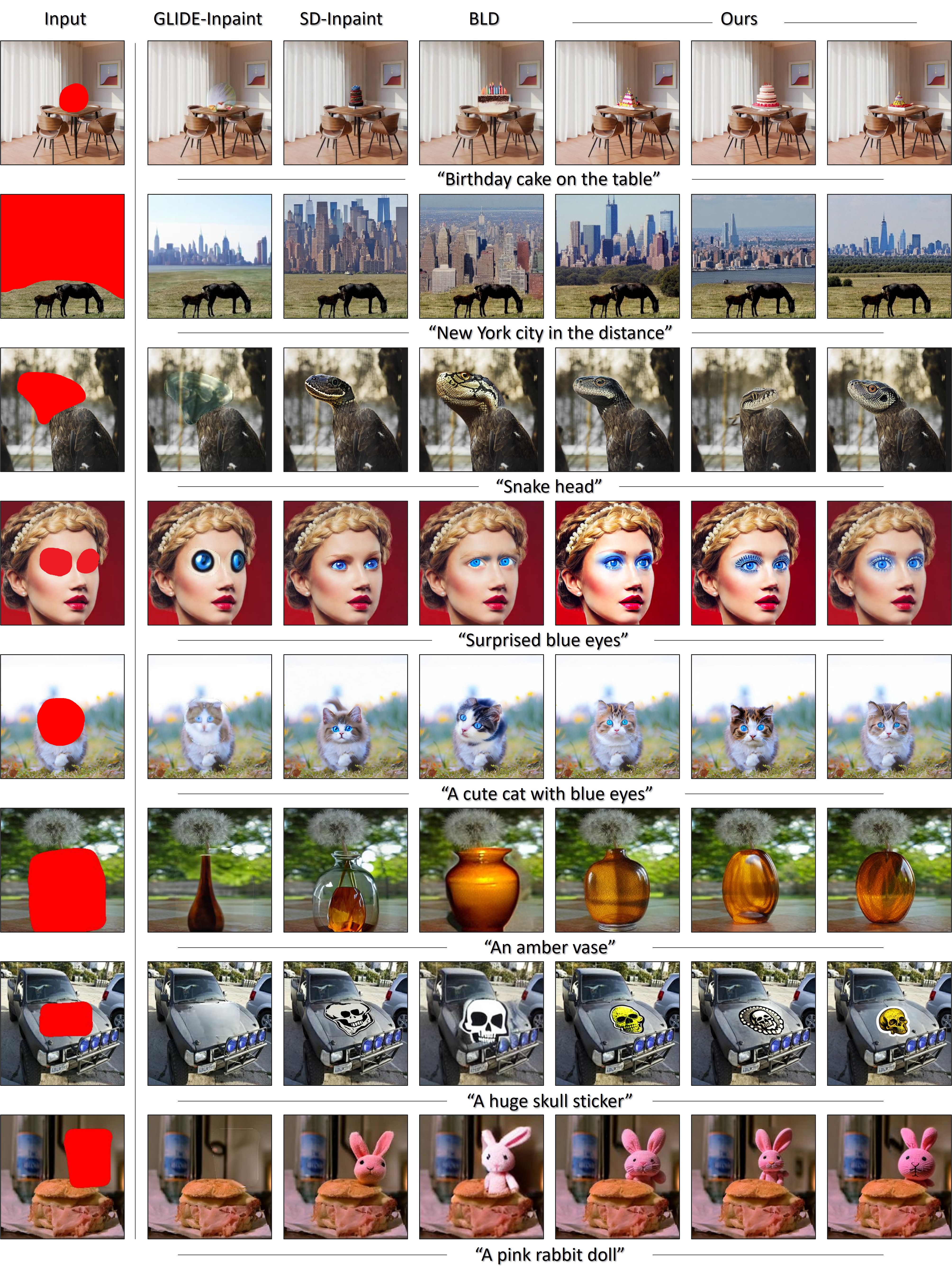}
\end{center}
   \caption{Additional text-driven inpainting results.}
\label{fig.apx.text}
\end{figure*}

\begin{figure*}[htpb]
\begin{center}
   \includegraphics[width=\linewidth]{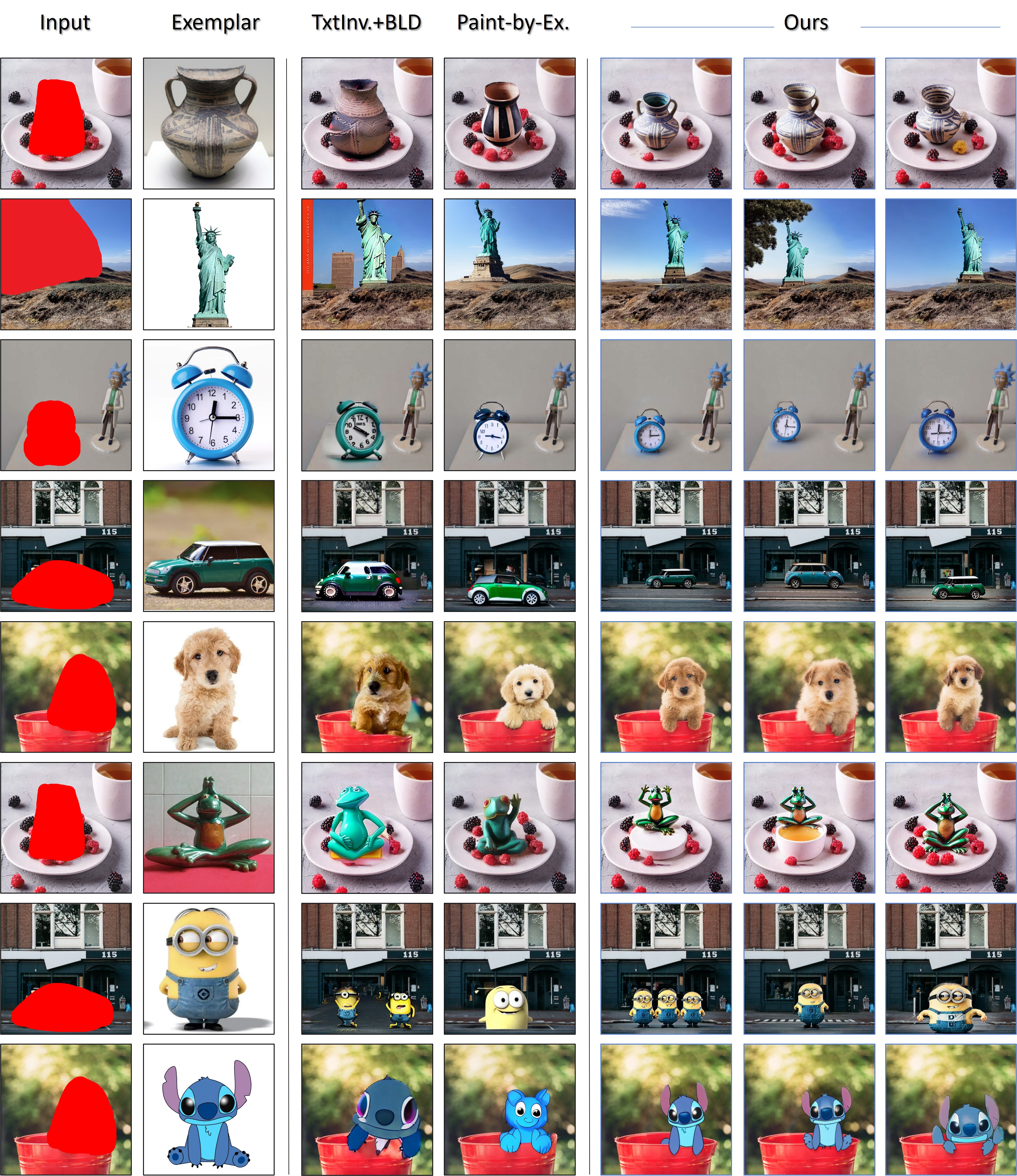}
\end{center}
   \caption{Additional exemplar-driven inpainting results.}
\label{fig.apx.exemplar}
\end{figure*}

\begin{figure*}[htpb]
\begin{center}
   \includegraphics[width=\linewidth]{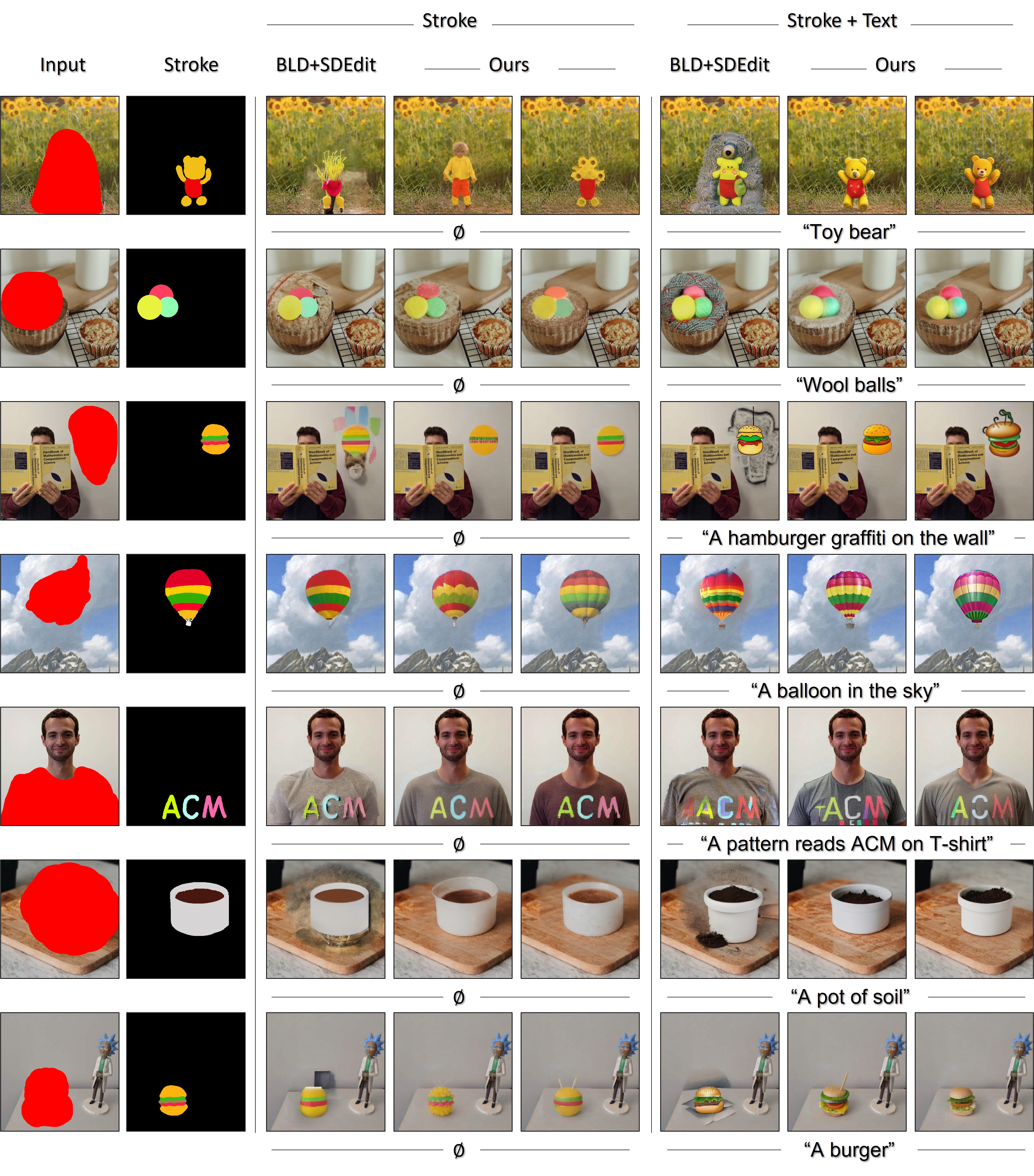}
\end{center}
   \caption{Additional stroke-driven inpainting results.}
\label{fig.apx.stroke}
\end{figure*}

\begin{figure*}[htpb]
\begin{center}
   \includegraphics[width=\linewidth]{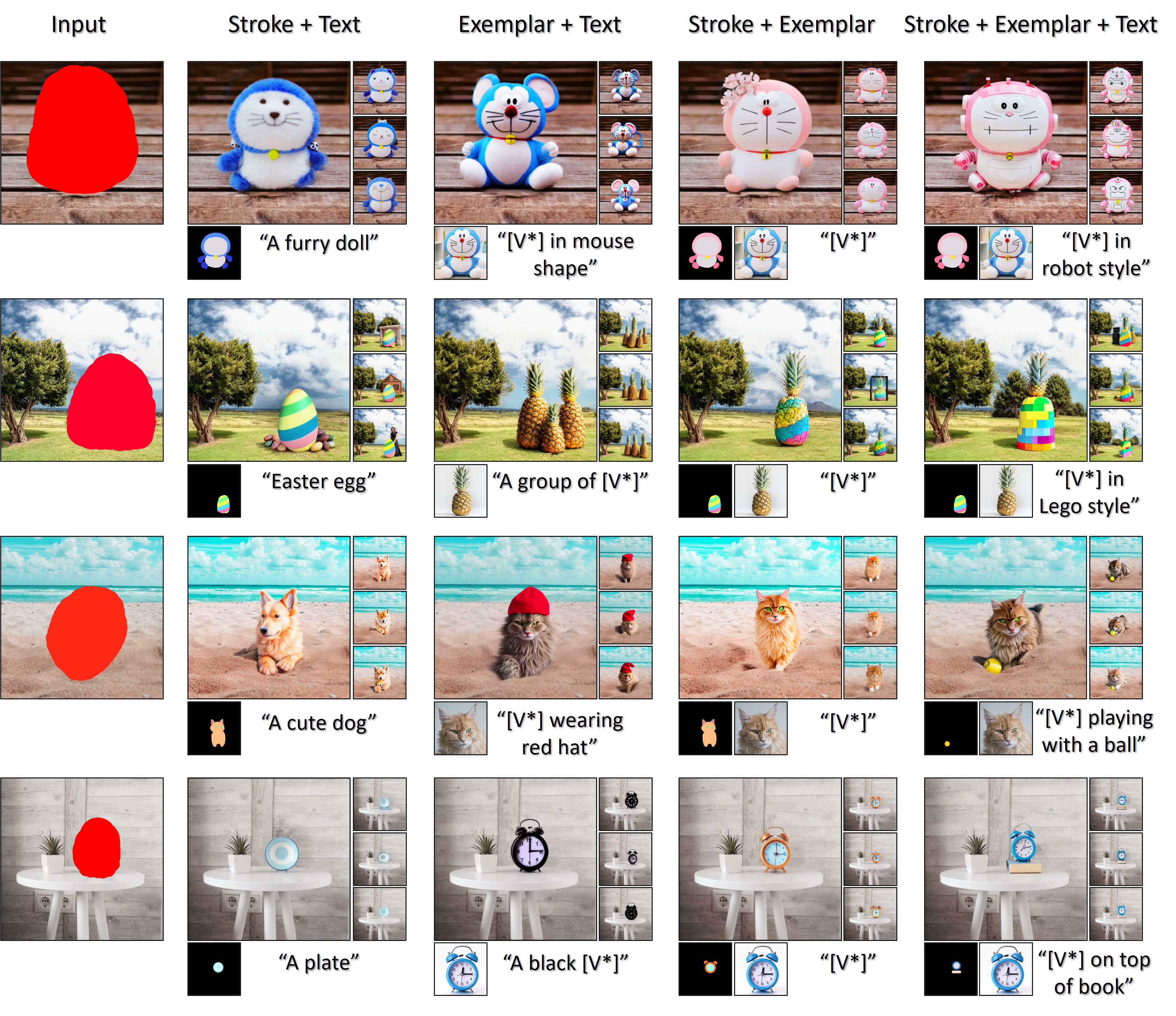}
\end{center}
   \caption{Additional inpainting results of mixed-guidance.}
\label{fig.apx.mixed}
\end{figure*}

We provide additional visual comparison with the same baseline methods as in the main paper. We show more unconditional, text-driven, exemplar-driven and stroke-driven result comparison in Fig.~\ref{fig.apx.uncond}, \ref{fig.apx.text}, \ref{fig.apx.exemplar}, and \ref{fig.apx.stroke}, respectively. We also present additional results of mixed guidance in Fig.~\ref{fig.apx.mixed}.

%%%%%%%%%%%%%%%%%%%%%%%%%%%%%%%%%%%%%%%%%%%%%%%%%%%%%%%%%%%
% \section{Demo}\label{apx.demo}
% We designed an interactive tool for our method, Fig.~\ref{fig.apx.ui} shows a screenshot of our tool. Please refer to our attached video for more detailed demonstration.

% \begin{figure*}[htpb]
% \begin{center}
%    \includegraphics[width=\linewidth]{fig.apx.ui.jpg}
% \end{center}
%    \caption{Screenshot of our demo interface.}
% \label{fig.apx.ui}
% \end{figure*}

\end{document}